\documentclass[runningheads,envcountsect,envcountsame,orivec]{llncs}
\usepackage{iftex}
\usepackage[utf8]{inputenc}
\usepackage[T1]{fontenc}
\usepackage{graphicx}

\usepackage{amsmath}

\usepackage{amsthm}

\usepackage{amssymb}
\usepackage{graphicx}
\usepackage{float}
\usepackage{enumerate}
\usepackage{multicol}
\usepackage{csquotes}
\usepackage{xcolor}
\usepackage{subcaption}
\usepackage{xparse}
\usepackage{tikz}
\usepackage{url}
\usepackage{array}
\usepackage{etoolbox}
\usepackage[hidelinks]{hyperref}
\hypersetup{hidelinks}

\usepackage{graphicx}
\usepackage{xspace}
\usepackage{xparse}
\usepackage{enumerate}
\usepackage{tabularx}
\usepackage{booktabs}
\usepackage{csquotes}
\usepackage{array}
\usepackage{latexsym}
\usepackage{wasysym}
\usepackage{multicol}
\usepackage{multirow}
\usepackage{subcaption}
\usepackage{arydshln}
\usepackage{rotating}
\usepackage{stmaryrd}
\usepackage{todonotes}
\usepackage{microtype}

\usepackage{tikz}
\usetikzlibrary{fadings}
\usetikzlibrary{shapes,decorations,positioning}
\usetikzlibrary{fadings,shapes,decorations,positioning,calc}
\usetikzlibrary{decorations.markings,decorations.pathreplacing}
\usetikzlibrary{shapes.geometric,automata,positioning}
\usetikzlibrary{shadows}\usetikzlibrary{calc}
\usetikzlibrary{decorations,decorations.pathmorphing,decorations.pathreplacing}
\usetikzlibrary{calc}

\newcolumntype{L}[1]{>{\raggedright\let\newline\\\arraybackslash\hspace{0pt}}m{#1}}
\newcolumntype{C}[1]{>{\centering\let\newline\\\arraybackslash\hspace{0pt}}m{#1}}
\newcolumntype{R}[1]{>{\raggedleft\let\newline\\\arraybackslash\hspace{0pt}}m{#1}}

\usepackage[hidelinks]{hyperref}
\makeatletter
\newcommand{\ifLatexThree}[2]{\@ifpackageloaded{xparse}{#1}{#2}}
\newcommand{\ifAMSmath}[2]{\@ifpackageloaded{amsmath}{#1}{#2}}
\newcommand{\ifMathSCR}[2]{\@ifpackageloaded{mathrsfs}{#1}{#2}}
\newcommand{\ifMathHyperREF}[2]{\@ifpackageloaded{hyperref}{#1}{#2}}
\makeatother

\ifLatexThree{%
	\NewDocumentCommand{\headword}{s o m}{\IfBooleanTF{#1}{#3}{\textbf{#3}}\IfNoValueTF{#2}{\index{#3}}{\index{#2}}}%
}{%
	\makeatletter
	\def\@headword#1{\textbf{#1}\index{#1}}%
	\def\@@headword#1{#1\index{#1}}%
	\def\headword#1{\@ifstar\@headword{#1}\@@headword{#1}}%
	\makeatother
}

\makeatletter
\@ifundefined{naturals}{}{}
\@ifundefined{ol}{}{}
\makeatother

\makeatletter
\newcommand{\textlabelmarker}[1]{%
	\protected@edef\@currentlabel{#1}%
	\phantomsection%
}
\newcommand{\textlabel}[2]{%
	\textlabelmarker{#1}%
	#1\label{#2}%
}
\makeatother

\ifx\nmableit\undefined
\makeatletter
\ifx\centernot\undefined
\newcommand*{\centernot}{%
	\mathpalette\@centernot
}
\fi
\def\@centernot#1#2{%
	\mathrel{%
		\rlap{%
			\settowidth\dimen@{$\m@th#1{#2}$}%
			\kern.5\dimen@
			\settowidth\dimen@{$\m@th#1=$}%
			\kern-.5\dimen@
			$\m@th#1\not$%
		}%
		{#2}%
	}%
}
\makeatother
\DeclareRobustCommand\nmableitSymb{\mathrel|\mkern-.5mu\joinrel\sim} %
\newcommand{\nmableit}{\ensuremath{\mbox{$\,\nmableitSymb\,$}}} %
\fi

\makeatletter
\ifMathHyperREF{%
	\newcommand{\seqref}[1]{\hyperref[{#1}]{\textup{\tagform@split{\getrefnumber{#1}}}}}%
}{%
	\newcommand{\seqref}[1]{\textup{\tagform@split{\getrefnumber{#1}}}}%
}
\newcommand\tagform@split[1]{%
	\begingroup
	\m@th\normalfont(\ignorespaces #1\unskip\@@italiccorr)%
	\endgroup
}
\makeatother

	\makeatletter
	\newcommand{\leqnomode}{\tagsleft@true\let\veqno\@@leqno}
	\newcommand{\reqnomode}{\tagsleft@false\let\veqno\@@eqno}
	\newcommand{\pushright}[1]{\ifmeasuring@#1\else\omit\hfill$\displaystyle#1$\fi\ignorespaces}
	\newcommand{\pushleft}[1]{\ifmeasuring@#1\else\omit$\displaystyle#1$\hfill\fi\ignorespaces}
	\newcommand{\specialcell}[1]{\ifmeasuring@#1\else\omit$\displaystyle#1$\ignorespaces\fi}

	\makeatother
\newcommand{\ksIF}{\text{if }}
\newcommand{\ksTHEN}{\text{, then }}

\newcommand{\ksAND}{\text{ and }}
\newcommand{\ksOR}{\text{ or }}

\newcommand{\ksOtherwise}{\text{otherwise}}

\newcommand{\tuple}[1]{\ensuremath{\langle{#1}\rangle}}

\newcommand{\minOf}[2]{\ensuremath{\min(#1,#2)}}

\ifMathSCR{%
}{%
}

\makeatletter
\newcommand\footnoteref[1]{\protected@xdef\@thefnmark{\ref{#1}}\@footnotemark}
\makeatother

\usepackage{environ,xparse,xkeyval,ifthen}
\newif\ifpostulatepresent\postulatepresentfalse

\ExplSyntaxOn
\NewEnviron{postulate}[2]%
{%
	\tl_gclear:N \g_tmpa_tl%
	\tl_gput_right:Nn \g_tmpa_tl { \ifpostulatepresent\\[\smallskipamount]\fi\global\postulatepresenttrue }%
	\tl_gput_right:Nn \g_tmpa_tl { \noindent\ignorespaces\ifthenelse{\equal{#1}{}}{}{(\textlabel{#1}{pstl:#1})} & }%
	\tl_gput_right:No \g_tmpa_tl { \BODY }%
	\tl_gput_right:Nn \g_tmpa_tl { & \hspace{1.5\medskipamount} #2  }%
	\group_insert_after:N \g_tmpa_tl%
}
\ExplSyntaxOff

\DeclareMathOperator{\supmin}{\supmin}

\newcommand{\ksChoiceFunction}{\ensuremath{\triangledown}}
\newcommand{\ksChoiceImage}[1]{\ensuremath{\textsf{Image}({#1})}}
\newcommand{\ksChoiceOrderEF}{\ensuremath{\unlhd}}

\makeatletter
\renewcommand{\textlabel}[2]{%
    \protected@edef\@currentlabel{#1}%
    \phantomsection%
    #1\label{#2}%
}
\makeatother

\newcommand{\mins}{\operatorname{mins}}
\newcommand{\minsOf}[2]{\ensuremath{\mins(#1,#2)}}
\DeclareMathAlphabet{\altmathcal}{OMS}{cmsy}{m}{n}
\newcommand{\ksPowerset}[1]{\ensuremath{\altmathcal{P}(#1)}}
\newcommand{\ksMathcal}[1]{\ensuremath{\altmathcal{#1}}}
\newcommand{\ksLogicLang}{\ensuremath{\altmathcal{L}}}

\DeclareFontFamily{U}{mathb}{\hyphenchar\font45}
\DeclareFontShape{U}{mathb}{m}{n}{%
    <-6> mathb5
    <6-7> mathb6
    <7-8> mathb7
    <8-9> mathb8
    <9-10> mathb9
    <10-12> mathb10
    <12-> mathb12 }{}
\DeclareSymbolFont{mathb}{U}{mathb}{m}{n}
\DeclareMathSymbol{\sqbullet}{\mathbin}{mathb}{"0D}
\DeclareMathSymbol{\sqsquare}{\mathbin}{mathb}{"05}

\begin{document}
\title{Axiomatics of Restricted Choices by
        Linear~Orders~of~Sets with Minimum as Fallback}
        \subtitle{(Including Supplemental Material)}
\titlerunning{Axiomatics of Restricted Choices by~Linear~Orders of Sets}
\author{Kai Sauerwald\inst{1}%
    \and
    Kenneth Skiba\inst{1}%
    \and
    Eduardo Fermé\inst{2}%
    \and
    Thomas~Meyer\inst{3}%
    }
\authorrunning{Sauerwald, Skiba, Fermé, and Meyer}
\institute{University of Hagen, Germany
    \and
    University of Madeira and NOVA-LINCS, Portugal
    \and
    University of Cape Town and CAIR, South Africa}
\maketitle              %
\setcounter{footnote}{0} 
\begin{abstract}
We study how linear orders can be employed to realise choice functions for which the set of potential choices is restricted, i.e., the possible choice is not possible among the full powerset of all alternatives. In such restricted settings, constructing a choice function via a relation on the alternatives is not always possible. However, we show that one can always construct a choice function via a linear order on sets of alternatives, even when a fallback value is encoded as the minimal element in the linear order. The axiomatics of such choice functions are presented for the general case and the case of union-closed input restrictions. Restricted choice structures have applications in knowledge representation and reasoning, and here we discuss their applications for theory change and abstract argumentation.     
\end{abstract}

\section{Introduction}
\label{sec:introduction}
\newcommand{\ksChoiceDomain}{\ensuremath{\mathbb{S}}}
\newcommand{\ksChoiceCoDomain}{\ensuremath{\mathbb{E}}}
\newcommand{\ksChoiceAlternatives}{\ensuremath{A}}
\newcommand{\ksChoiceStructure}{\ensuremath{\ksMathcal{R}}}

A (classical) choice structure \( \tuple{\ksChoiceAlternatives,\ksChoiceDomain} \) consists of a set of alternatives \( \ksChoiceAlternatives \) and a set of subsets \( \ksChoiceDomain \subseteq \ksPowerset{\ksChoiceAlternatives} \) of \( \ksChoiceAlternatives \).
A (classical) choice function \( O: \ksChoiceDomain \to \ksPowerset{\ksChoiceAlternatives} \) for \( \tuple{\ksChoiceAlternatives,\ksChoiceDomain} \)  maps each set \( S \in \ksChoiceDomain \) to a subset of \( S \).
Some authors also demand that \( O(S) \) is non-empty if and only if \( S \) is non-empty and in some communities, choice functions are required to output a singleton whenever \( S \) is non-empty.

\begin{example}\label{ex:snacks1}
	We are planning to buy snacks in a supermarket for the evening. 
	From experience, one knows that the typical snacks that are available in a supermarket are \( \ksChoiceAlternatives=\{ \texttt{chocolate},\allowbreak \texttt{nachos},\allowbreak \texttt{pretzels},\allowbreak \texttt{dips},\allowbreak \texttt{chillies} \} \). We determine that we want to buy only salty snacks \( S= \{ \allowbreak \texttt{nachos},\allowbreak \texttt{pretzels},\allowbreak \texttt{dips} \} \). Because one does not want to buy all available options, one might decide to choose from one of the available options \( O(S)=\{  \texttt{pretzels}, \texttt{nachos} \} \).
    \hfill\( \blacksquare \)
\end{example}

\noindent Work on choices has applications in, e.g., social choice theory and economics~\cite{KS_Sen1971}, mathematics and logic~\cite{KS_Zermelo1904}, and knowledge representation (KR)~\cite{KS_Rott2001,KS_Haret2020}. 
Here, we are aiming at its impact on KR, where approaches to choice are connected to nonmonotonic reasoning, belief change, update, and conditionals~\cite{KS_Makinson1993}. 
Research in semantics of, e.g., belief revision, shows that revisions satisfy desired properties exactly when the result of revision is chosen by employing an underlying preference structure~\cite{KS_KatsunoMendelzon1992}.
However, as observed, the connection between known axiomatizations and choice developed, which hold, e.g., for propositional logic, does not carry over to certain settings.
 Examples are preferential team-based logics, where the axiomatization turns out to be difficult~\cite{KS_SauerwaldKontinen2024}, Horn logic~\cite{KS_CreignouPapiniPichlerWoltran2014,KS_DelgrandePeppas2015}, or belief change in epistemic spaces, where belief change operators are not realizable in certain circumstances~\cite{KS_SauerwaldThimm2024,KS_Sauerwald2024}.
The rationale why classical connections do not carry over to these settings is that the output of potential choices is restricted in these settings. 
The choice function cannot simply output an element of the full powerset over the alternatives, i.e., \( O(S) \) \enquote{should} be a certain set, but \( O(S) \) is not expressible, e.g. in Horn logic.
The following 
Example~\ref{ex:snacks2a} illustrates that one is often more restricted in choice on the output side than the original setting of choice permits, i.e., the codomain of a choice \( O(S) \) is not the full powerset of all alternatives.
\begin{example}[Continued from Example~\ref{ex:snacks1}]\label{ex:snacks2a}
	Originally, in Example~\ref{ex:snacks1}, we decided to buy \( \texttt{pretzels} \) and \( \texttt{nachos} \), but nothing else.
	However, the supermarket we are visiting offers \( \texttt{nachos} \), \( \texttt{dips} \) and \( \texttt{chillies} \) only in a bundle together. 
	This means that the choice \( O(S)=\{  \texttt{pretzels}, \texttt{nachos} \} \) given in Example~\ref{ex:snacks1} is not valid in this setting, because the supermarket has no such offer in their stock.
    \hfill\( \blacksquare \)
\end{example}

Existence and non-existence of choice functions in the classical setting of choice has been studied extensively~\cite{KS_BossertSuzumura2010}. 
The existence of any choice functions is not guaranteed when considering arbitrary sets within the Zermelo–Fraenkel (ZF) set theory~\cite{KS_Jech2002}.
This can be resolved by adding the Axiom of Choice (ZFC), which enforces the existence of choice functions (that output only singletons):
\begin{center}
	\begin{minipage}[t]{0.95\columnwidth}
		{(Axiom of Choice) } \begin{minipage}[t]{0.75\linewidth}
			For each set of non-empty sets \( \ksChoiceDomain \), there exists a function \( f: \ksChoiceDomain \to \bigcup_{S\in\ksChoiceDomain} S \) with \( f(S)\in S \) for every \( S\in\ksChoiceDomain \).
		\end{minipage} 
	\end{minipage}
\end{center}
From social choice theory~\cite{KS_Kelly2013}, it is well-known that already mild assumptions about the representation of choice functions lead to the non-existence of choice functions, even when one assumes ZFC.
An often considered case is the construction of choice functions via a relation. This is  done by considering a preference relation \( {\leq} \subseteq \ksChoiceAlternatives\times\ksChoiceAlternatives \) on the alternatives \( \ksChoiceAlternatives \). A choice function \( O \) is then defined by letting \( O(S) \) be the most preferred elements within \( S \) according to \( {\leq} \), which we write as \( O(S)=\minOf{S}{{\leq}} \).\footnote{We follow here the convention that being smaller in \( \leq \) corresponds to being more preferred, which is in line with the typical reading of orders in belief revision and nonmonotonic reasoning. Technically, one could also consider everything from a dual perspective, where the larger elements are the more preferred elements.}  
Many authors follow the suggestion from Sen's seminal work~\cite{KS_Sen1971} to consider unrestricted inputs, i.e., having \( \ksChoiceDomain = \ksPowerset{\ksChoiceAlternatives} \). 
Effectively, many theoretical results for choice functions are given under this condition.
In this paper, we study in a general manner the setting where choices are limited to a set of choices \( \ksChoiceCoDomain \subseteq \ksPowerset{\ksChoiceAlternatives} \) (which we call the realizable choices) and where the sets to choose from are limited to \( \ksChoiceDomain \subseteq \ksPowerset{\ksChoiceAlternatives} \).
A \emph{restricted choice function} is a function \( C: \ksChoiceDomain \to \ksChoiceCoDomain \) where \( C(S)\subseteq S \) holds, whenever possible. 
Formally, the setting of restricted choice functions enables us to model the situation from Example~\ref{ex:snacks2a} easily.
\begin{example}[Continued from Example~\ref{ex:snacks2a}]\label{ex:snacks2}
We are only permitting choices where \( \texttt{nachos} \), \( \texttt{dips} \) and \( \texttt{chillies} \) appear together, i.e., by setting \( \ksChoiceCoDomain \) to
 \begin{align*}
 	\ksChoiceCoDomain & = \{ X \subseteq \ksChoiceAlternatives  \mid  \begin{aligned}[t]
         & (\{  \texttt{nachos},\texttt{dips},\texttt{chillies} \} \cap X \neq \emptyset)\\
         & \text{ implies } \{  \texttt{nachos},\texttt{dips},\texttt{chillies} \} \subseteq X \}
     \end{aligned} \\
 	& = \big\{ 
 	 \{ \texttt{pretzels} \},\ 
 	 \{ \texttt{chocolate} \},\ 
 	 \{ \texttt{pretzels} , \texttt{chocolate} \},\\
      & \hspace{0.7cm} \{ \texttt{nachos},\texttt{dips}, \texttt{chillies} \},\ \{ \texttt{pretzels}, \texttt{nachos},\texttt{dips}, \texttt{chillies} \},\\
      &\hspace{0.7cm} \{ \texttt{chocolate}, \texttt{nachos},\texttt{dips}, \texttt{chillies} \},\
 	\ksChoiceAlternatives
 	\big\} \ .
 \end{align*}
Consequently, the only valid output \( C(S) \) for \( S= \{ \texttt{pretzels},\allowbreak \texttt{nachos},\allowbreak \texttt{dips} \} \) from Example~\ref{ex:snacks1} is 
\( C(S)=\{ \texttt{pretzels} \} \) because we decided beforehand not to buy chillies, i.e., \( \texttt{chillies} \notin S \). 
Moreover, we have thought carefully about snacks and come to the conclusion that we only want \( \texttt{nachos} \) when we can also get \( \texttt{dips} \).
One can model such personal restrictions on the input side, e.g., by considering restricted choice functions for \( \ksChoiceDomain = \{ S \subseteq X  \mid  \texttt{nachos} \in S \Rightarrow \texttt{dips} \in S \} \).
\hfill\( \blacksquare \)
\end{example}
\smallskip 

The setting of restricted choice is a powerful generalization of the classical setting that permits specification of restrictions on the output side. 
Two basic conceptual problems arise, which we resolve in this paper:
\begin{itemize}
	\item Unrealisable choices
	\item Existence and representation of restricted choice functions
\end{itemize}
The latter is because the axiom of choice does not immediately guarantee the existence of a restricted choice function. Specifically, it is not clear how a choice function for restricted choice structures can be represented or constructed via orders. Here, we show that an efficient way is to employ a linear order on sets of alternatives.
The problem of unrealizable choices arises because there are situations where it is impossible to yield a choice for a set \( S\in\ksChoiceDomain \) because there is no \( E \in\ksChoiceCoDomain \) in the co-domain of \( C \) such that \( E \subseteq S \).
However, as \( C \) is a function, there has to be an output for \( C(S) \).
\begin{example}
	Consider again the situation discussed in Example~\ref{ex:snacks2} where \texttt{nachos} are bundled with \texttt{dips} and \texttt{chillies}. 
	For example, for \( S' = \{ \texttt{nachos}, \texttt{dips}  \} \) there is no  \( E \in\ksChoiceCoDomain \) such that \( E \subseteq S' \).
\end{example}
\noindent Here, we solve the problem of unrealisable choices by outputting a fallback value. For that, we encode the fallback value as the minimal element in a linear order.

We demonstrate how restricted choice functions can be employed in theory change and abstract argumentation. 
	Specifically, we show that our representation theorem for linear choice functions carries over to theory change and abstract argumentation.
	We see that linear choice functions lead to a natural general approach to theory change operators. 
	In abstract argumentation, we show that employing choice functions leads to a new approach to argumentation semantics, which has not been explored and generalizes extension selection~\cite{DBLP:conf/ecsqaru/KoniecznyMV15,DBLP:conf/kr/BonzonDKM18,KS_SkibaRienstraThimmHeyninckKernIsberner2021}.

\pagebreak[3]
In summary, the main content of the paper is the following, which we also consider as the main contributions of this paper\footnote{This paper is a preprint of a paper presented at JELIA 2025~\cite{KS_SauerwaldSkibaFermeMeyer2025}, extended by proofs.\label{ftn:arxiv}}:
\begin{itemize}
	\item {{[Restricted Choice Structures and their Choice Functions]}} We introduce the concepts of restricted choice structures and (restricted) choice functions for restricted choice structures. 
	Restricted choice structures provide a more expressive setting than the original choice setting that permits taking unrealisable choices into account.
	
	\item {[{Linear Choice Functions}]} We present a uniform way of constructing restricted choice functions by employing a linear order on the available outcomes. We show that this construction provides a restricted choice function for any restricted choice structure, a feature that typical construction methods via orders on alternatives do not guarantee. Moreover, the full axiomatization of linear order-based choice functions is provided for both union-closed restricted choice structures and arbitrary restricted choice structures.
	\item {[{Applications in KR}]} We discuss the application of restricted choice structures and their choice functions in knowledge representation and reasoning (KR).
	We discuss the application in theory revision and argumentation.
\end{itemize}
In the next section, we start by presenting the background of order theory.
In Section~\ref{sec:RCS} we formally introduce restricted choice structures and their choice functions.
Linear choice functions are introduced in Section~\ref{sec:LCS}.
The axiomatics of linear choice functions is given in Section~\ref{sec:AxiomaticsKMinimal}.
Section~\ref{sec:applyKR} is dedicated to explore applications of the restricted choice structures and linear choice functions in knowledge representation and reasoning;
namely theory change (Section~\ref{sec:applyTC}) and argumentation (Section~\ref{sec:applyArg}).
The conclusion of this paper is given in Section~\ref{sec:conclusion}.

\section{Background on Relations and Order Theory}
\label{sec:background}
We use \( \mathbb{N} \) for the natural numbers including \( 0 \), and \( \mathbb{N}^+ \) for the natural numbers excluding~\( 0 \). 
The powerset of a set \( X \) is denoted by \( \ksPowerset{X} \).
In the following, we present the background on basic notions of relations and order theory. Moreover, we present some basic background in extensions of relations to orders.

\medskip
\noindent\textbf{Relations and Orders.}
A (binary) relation on a set \( X \) is a subset of \( X\times X \). 
We will often use order symbols for relations and write them infix, e.g., \( x \preceq y \) means the same as \( (x,y)\in {\preceq} \) for a relation \( {\preceq} \).
With \( \prec \) we denote the strict part of a relation \( \preceq \) on \( X \) and with \( {\simeq} \) we denote the equivalent part of \( {\preceq} \), i.e., \( {\simeq}  = {\preceq} \cap \{ (x_2,x_1) \in X \times X \mid x_1 \preceq x_2 \} \) and \( {\prec} =  {\preceq} \setminus {\simeq} \).
The following properties of a relation \( {\preceq} \subseteq X \times X \) are considered in this article:

\medskip\noindent
\begin{tabular}{@{}l@{\ }l@{}}
    (\textlabel{reflexive}{pstl:reflexive}) & \text{\( x \preceq x \)} \\[0.25em]
    (\textlabel{total}{pstl:total}) & \text{\( x_1 \preceq x_2\)} \\[0.25em]
    (\textlabel{antisymmetric}{pstl:antisymmetric}) & \text{\( x_1 \preceq x_2 \) and \( x_2 \preceq x_1 \) imply \( x_1 = x_2 \)}\\[0.25em]
    (\textlabel{transitive}{pstl:transitive}) & \text{\( x_1 \preceq x_2 \) and \( x_2 \preceq x_3 \) imply \( x_1 \preceq x_3 \)}\\[0.25em]
    (\textlabel{consistent}{pstl:Suzumura_consistency}) & \text{\( x_0 \preceq x_1 \), ..., \( x_{n-1} \preceq x_{n} \) implies \( x_{n} \not\prec x_{0} \)}		
\end{tabular}
\medskip

\noindent 
A \emph{preorder} \( {\preceq} \) on a set \( X \) is a relation \( {\preceq}\subseteq X \times X \) such that \( {\preceq} \) is reflexive and transitive. We also consider \emph{total preorders}, i.e., preorders that also satisfy totality. 
A \emph{linear order} on a set \( X \) is a total preorder on \( X \) that is additionally antisymmetric.
The \hyperref[pstl:Suzumura_consistency]{consistency} property is due to Suzumura and has a central place in order theory, as it guarantees the existence of an order-extension~\cite{KS_Suzumura1976}.
A subtle aspect of the \hyperref[pstl:Suzumura_consistency]{consistency} property is that one demands \( x_{n} \not\prec x_{0} \) instead of  \( x_{n} \not\preceq x_{0} \).
This ensures that the \hyperref[pstl:Suzumura_consistency]{consistency} property does not apply when \( x_{n} \simeq x_{0} \) holds.
If \( \preceq \) is transitive, then  \( \preceq \) is also consistent, and if \( \preceq \) is total, then \( \preceq \) is also reflexive.
\medskip
\noindent\textbf{Minimal Elements.}
We define two types of minimal elements for \( {\preceq} \subseteq X \times X \), the (globally) minimal elements \( \min({\preceq}) \), and the minimal elements \( \minOf{M}{\preceq} \) with respect to a set \( M  \), which are given by:
\begin{equation*}
    \begin{array}{rl@{}l}
        \min(\preceq) & =\{\  x \in X \ &\mid x' \preceq x \text{ implies } x \preceq x' \text{ for all } x'\in X \}\\
        \minOf{M}{\preceq} & =\{\  x \in M\cap X \ &\mid  x' \preceq x \text{ implies } x \preceq x' \text{ for all } x'\in M \}
    \end{array}
\end{equation*}
Note that for arbitrary relations (even on a finite set) there might be no minimal elements, i.e., \( \min(\preceq) \) and \( \minOf{M}{\preceq} \) might be empty sets.
However, if \( \preceq \) is a total preorder on a finite (non-empty) set, then \( \min(\preceq) \) is always non-empty, and \( \minOf{M}{\preceq} \) is only empty if \( M \) is empty.
We say that \( {\preceq} \subseteq X \times X \) is \emph{well-founded} if for each \( M \subseteq X \) holds \( \minOf{M}{\preceq}\neq\emptyset \).

We deal in this paper with relations on sets of sets \( \ksMathcal{E} \subseteq \ksPowerset{A} \) over some base-set \( A \).
For such a relation \( {\leq} \subseteq \ksMathcal{E} \times \ksMathcal{E} \), we overload the notion of  \( \minOf{\cdot}{\cdot} \) by extending it to elements from \( \ksPowerset{A} \).
When \( S \in \ksPowerset{A} \) is an element from \( \ksPowerset{A} \), then we let
    \( \minOf{S}{\leq} = \minOf{\, \{ E \in \ksMathcal{E} \mid E \subseteq S \} \, }{\leq}  . \)
That is, \( \minOf{S}{\leq} \) is the set of \( \leq \)-minimal elements among all elements from \( \ksMathcal{E} \) that are also subsets of \( S \).

\section{Restricted Choice Structures}
\label{sec:RCS}
We define restricted choice structures and their corresponding choice functions.

\begin{definition}\label{def:restricted_choice_structure}
	A (normal) restricted choice structure is a tuple \( \ksChoiceStructure = \tuple{\ksChoiceAlternatives,\ksChoiceDomain,\ksChoiceCoDomain} \) where
	\begin{itemize}
		\item \( \ksChoiceAlternatives \) is a set (the set of alternatives),
		\item \( \ksChoiceDomain \subseteq \ksPowerset{\ksChoiceAlternatives} \) is a non-empty  set of subsets of \( \ksChoiceAlternatives \) (the domain), and
		\item \( \ksChoiceCoDomain \subseteq \ksPowerset{\ksChoiceAlternatives} \) is a non-empty subset of \( \ksChoiceDomain \) (the realizable choices) with \( \ksChoiceCoDomain \subseteq \ksChoiceDomain \).
	\end{itemize}
We call \( \ksChoiceStructure \) \emph{(input) union-closed} if \( S_1 \cup S_2 \in \ksChoiceDomain \) holds for all \( S_1,S_2 \in \ksChoiceDomain \).
\end{definition}
Next, we define the notion of choice functions for restricted choice structures.

\begin{definition}
    Let \( \ksChoiceStructure = \tuple{\ksChoiceAlternatives,\ksChoiceDomain,\ksChoiceCoDomain} \) be a restricted choice structure. 
    A function \( C: \ksChoiceDomain \to \ksChoiceCoDomain \) is called a \emph{choice function for \( \ksChoiceStructure \)} if for each \( S \in \ksChoiceDomain \) holds \( C(S) = E \in \ksChoiceCoDomain  \) with \( E \subseteq S \), if such an \( E \in \ksChoiceCoDomain\) exists.
    If \( C(S)=K \) for all \( S \in \ksChoiceDomain \) for which no \( E \in \ksChoiceCoDomain \) with \( E \subseteq S \) exists, we say that \( K \) is the \emph{fallback (value)} of~\( C \).
\end{definition}
Note that a (unrestricted) choice structure \( \tuple{A, \ksChoiceDomain} \) can \textbf{not} be simply reconstructed by taking \( \tuple{\ksChoiceAlternatives,\ksChoiceDomain,\ksPowerset{\ksPowerset{\ksChoiceAlternatives}}} \),\pagebreak[3] as the latter violates \( \ksChoiceCoDomain \subseteq \ksChoiceDomain \) and thus is not a (normal) restricted choice structures.
Instead, the restricted choice structure  \( \tuple{\ksChoiceAlternatives,\ksChoiceDomain,\ksChoiceCoDomain} \) with \( \ksChoiceCoDomain = \bigcup_{S \in \ksChoiceDomain} \ksPowerset{S} \) reconstructs \( \tuple{A, \ksChoiceDomain} \) in the following sense:
\begin{enumerate}[(a)]
    \item for every choice function \( O:  \ksChoiceDomain \to \ksPowerset{A} \) for \( \tuple{A, \ksChoiceDomain} \), there is a choice function \( C: \ksChoiceDomain \to \ksChoiceCoDomain \) for \( \tuple{\ksChoiceAlternatives,\ksChoiceDomain,\ksChoiceCoDomain} \), such that \( O(S) = C(S) \) for all \( S \in \ksChoiceDomain \).
    \item for every choice function \( C: \ksChoiceDomain \to \ksChoiceCoDomain \) for \( \tuple{\ksChoiceAlternatives,\ksChoiceDomain,\ksChoiceCoDomain} \), there is a choice function \( O:  \ksChoiceDomain \to \ksPowerset{A} \) for \( \tuple{A, \ksChoiceDomain} \), such that \( O(S) = C(S) \) for all \( S \in \ksChoiceDomain \).
\end{enumerate}

    One might get the impression that the last condition of \( \ksChoiceCoDomain \subseteq \ksChoiceDomain \) in Definition~\ref{def:restricted_choice_structure} is too restrictive,
    as one may imagine settings of restricted choice, where the condition \( \ksChoiceCoDomain \subseteq \ksChoiceDomain \) does not hold.
    However, without \( \ksChoiceCoDomain \subseteq \ksChoiceDomain \), we have to deal with (potentially) unwanted consequences for choice functions \( C \). Examples are:
    \begin{itemize}
        \item \( \ksChoiceCoDomain \) contains elements that are not in the image of any choice function at all;
        \item choice might be not chainable, e.g., \( C(C(S)) \) is undefined;
        \item \( C(K) \) is undefined, while \( K \) is the fallback of \( C \).
    \end{itemize}
    Because of that, we call a restricted choice structure without \( \ksChoiceCoDomain \subseteq \ksChoiceDomain \) \emph{non-normal}.
    In this paper, we refrain from considering non-normal restricted choice structures and assume normality for the remainder of the paper.
    However, non-normal restricted choice structures might be of interest in future work.

\section{Linear Choice Functions}
\label{sec:LCS}
In this section, we show how to construct a choice function for a restricted choice structure by employing a linear order on the sets of realizable choice sets.
Not all linear orders are suitable for the approach.
The property we require is the existence of minima for all cases in which one wants to make choices. We call this property \emph{smoothness}. Fallbacks will be encoded as the globally minimal element.
\begin{definition}[\( \ksChoiceStructure \)-smoothness, \( K \)-minimal]
\label{def:smoothness}
    Let \( \ksChoiceStructure = \tuple{\ksChoiceAlternatives,\ksChoiceDomain,\ksChoiceCoDomain} \) be a restricted choice structure, let \( K \in \ksChoiceCoDomain \)  and let \( {\leq} \subseteq \ksMathcal{E} \times \ksMathcal{E} \)  be a relation on a set \( \ksMathcal{E} \subseteq \ksChoiceCoDomain \).
    We say that \( {\leq} \) is \emph{\( \ksChoiceStructure \)-smooth} if for each \( S \in \ksChoiceDomain \) holds \( \minOf{S}{\leq} \neq \emptyset \) whenever there is some \( E \in \ksMathcal{E} \) such that \( E \subseteq S \).
    We say that \( {\leq} \) is \emph{\(K\)-minimal} if \( \min(\leq) = \{K\} \).
\end{definition}

One might compare the notion of smoothness with the notion of a well-founded relation.
The difference is that one demands the existence of minimal elements only for certain elements of interest, which is a more liberal requirement.
The closest related notion is the notion of smoothness by Kraus, Lehmann and Magidor \cite{KS_KrausLehmannMagidor1990}, for which it has been shown that one cannot replace smooth relations with well-founded relations \cite{KS_LehmannMagidor1992}.
\(K\)-minimality means that \( K \) is the (globally unique) minimal element of \( {\leq} \).

For a \( \ksChoiceStructure \)-smooth relation, we define a corresponding choice function for \( \ksChoiceStructure \) and some \( K \), which will act as a fallback value.
Linear choice functions for \( \ksChoiceStructure \) are then such choice functions for \( \ksChoiceStructure \) given by a \( \ksChoiceStructure \)-smooth relation that is a linear order on \( \ksChoiceCoDomain \). If the relation is also \( K \)-minimal, we say that it is a \( K \)-minimal linear choice function for \( \ksChoiceStructure \), i.e., when the fallback value \( K \) is encoded as the (globally) minimal element.
\begin{definition}[linear choice function]\label{def:linear_choice_function}
	Let \( \ksChoiceStructure = \tuple{\ksChoiceAlternatives,\ksChoiceDomain,\ksChoiceCoDomain} \) be a restricted choice structure and let \( K \in  \ksChoiceCoDomain \).
	For each \( \ksChoiceStructure \)-smooth linear order \( {\ll} \subseteq \ksChoiceCoDomain \times \ksChoiceCoDomain \) on~\( \ksChoiceCoDomain \), we define the function \( \triangledown_{\ll}^{K}: \ksChoiceDomain \to \ksChoiceCoDomain \) as:
	\begin{equation}\tag{\( \star \)}\label{eq:triangledownLLdefinition}
		\triangledown_{\ll}^{K}(S) = \begin{cases}        
			E & \ksIF \minOf{S}{\ll}=\{ E \}\\
			K & \ksOtherwise
		\end{cases}
	\end{equation}%
	A function \( \triangledown: \ksChoiceDomain \to \ksChoiceCoDomain \) is a \emph{linear choice function for \( \ksChoiceStructure \)} if \( {\triangledown} = {\triangledown_{\ll}^{K}} \) for some \( \ksChoiceStructure \)-smooth linear order \( \ll \).
	Additionally, we say that \( {\triangledown_{\ll}^{K}} \) is \( K \)-minimal if \( \ll \)  is also \( K \)-minimal.
\end{definition}
Note that \( \ll \) in Definition~\ref{def:linear_choice_function} is a linear order on the full set~\( \ksChoiceCoDomain \). 
In contrast, the relation \( \leq \) in  Definition~\ref{def:smoothness} is a relation on any subset of~\( \ksChoiceCoDomain \).
The additional flexibility in the latter case is required for the proofs of theorems in Section~\ref{sec:AxiomaticsKMinimal}.

The following proposition witnesses that linear choice functions for a restricted choice structure \( \ksChoiceStructure \) are indeed choice functions for \( \ksChoiceStructure \). Moreover, when one assumes the Axiom of Choice, it is guaranteed that there exists some \( K \)-minimal linear choice function \( \ksChoiceStructure \).
\begin{proposition}\label{prop:linearchoice_existence}
    Let \( \ksChoiceStructure = \tuple{\ksChoiceAlternatives,\ksChoiceDomain,\ksChoiceCoDomain} \) be a restricted choice structure and let \( {K \in \ksChoiceCoDomain} \).
    The following statements hold:
    \begin{enumerate}[(a)]
    	\item \( \triangledown_{\ll}^{K}: \ksChoiceDomain \to \ksChoiceCoDomain \) is a  choice function for \( \ksChoiceStructure \) with fallback~\( K \) for any \( \ksChoiceStructure \)-smooth linear order \( {\ll} \subseteq \ksChoiceCoDomain \times \ksChoiceCoDomain \).
    	\item Assume the Axiom of Choice. There exists a \( K \)-minimal linear choice function for \( \ksChoiceStructure \).
    \end{enumerate}
\end{proposition}
\begin{proof}
	We start by showing Statement~(a).
	If some \( E \in \ksChoiceCoDomain \) with \( E \subseteq S \) exists, then \( \minOf{S}{\ll} = \{E'\} \) is non-empty and thus, according to \eqref{eq:triangledownLLdefinition}, we have \( \triangledown_{\ll}^{K}(S) = E'  \).
	This implies that we have \( \triangledown_{\ll}^{K}(S) \subseteq S  \). We obtain that \( \triangledown_{\ll}^{K} \) is a choice function for \( \ksChoiceStructure \). Moreover, if no \(  E \in \ksChoiceCoDomain \) with \( E \subseteq S \) exists, then we have \( \triangledown_{\ll}^{K}(S) = K \) due to  \eqref{eq:triangledownLLdefinition}. Consequently, \( K \)  is the fallback of \( \triangledown_{\ll}^{K} \).

	For Statement~(b), assume the Axiom of Choice. It is known that the Axiom of Choice is equivalent to the well-ordering theorem~\cite{KS_Jech2002}, which states that on every set \( S \) there is a linear order \( {\leq} \subseteq S \times S \) that is well-founded.
	Let \( {\ll'} \subseteq \ksChoiceCoDomain \times \ksChoiceCoDomain \) be such a well-founded linear order on \( \ksChoiceCoDomain \). Because \( {\ll'} \) is a well-founded linear order and \( \ksChoiceCoDomain \) is non-empty, there exist some unique minimal singleton set \( \{K'\}  = \minOf{\ksChoiceCoDomain}{\ll'}  \).
	Now, let \( {\ll} \subseteq \ksChoiceCoDomain \times \ksChoiceCoDomain \) be the linear order, in which \( K \) and \( K' \) are mutually substituted. This is the relation:
	\begin{align*}
		{\ll} = &\ \left({\ll'} \setminus \{ (K,S),(S,K),(K',S),(S,K') \mid S \in \ksChoiceCoDomain \} \right) \cup \{ (K,K), (K',K') \} \\
	 &	\hspace{1em} \cup \{ (K',S) \mid K \ll' S \ksAND S \neq K \} \cup \{ (S,K') \mid S \ll' K \ksAND S \neq K \}    \\
	 &	\hspace{1em} \cup \{ (K,S) \mid K' \ll' S \ksAND S \neq K' \} \cup \{ (S,K) \mid S \ll' K' \ksAND S \neq K' \}   \\
     &	\hspace{1em} \cup \{ (K',K) \mid K' \ll' K \}  \cup \{ (K,K') \mid K \ll' K' \}
	\end{align*}
	One can show that the relation \( {\ll} \) is a well-founded linear order on \( \ksChoiceCoDomain \). Clearly, we have \( {\minOf{\ksChoiceCoDomain}{\ll}} = \{  K \} \) and thus, \( \ll \) is \( K \)-minimal, and because \( \ll \) is well-wounded, \( \ll \) is \( \ksChoiceStructure \)-smooth. Statement~(b) follows by employing the latter observation and Statement~(a).
\end{proof}

\begin{example}[Continued from Example~\ref{ex:snacks2}]\label{ex:snacks3}
	We consider the restricted choice structure \( \ksChoiceStructure=\tuple{\ksChoiceAlternatives,\ksChoiceDomain,\ksChoiceCoDomain} \), where \(\ksChoiceAlternatives\), \(\ksChoiceDomain\) and \( \ksChoiceCoDomain\) are from Examples~\ref{ex:snacks1}--\ref{ex:snacks2}, and let \( K= \{ \texttt{pretzels}, \texttt{nachos},\texttt{dips}, \texttt{chillies} \} \).
	We let \( {\ll} \subseteq \ksChoiceCoDomain \times \ksChoiceCoDomain \) be the following linear order on \(  \ksChoiceCoDomain \):
	\begin{align*}
		& K \ll \{ \texttt{nachos},\texttt{dips}, \texttt{chillies} \}  \ll 	\{ \texttt{pretzels} , \texttt{chocolate} \} \ll	 \{ \texttt{pretzels} \} \\
		& \hspace{1cm}  \ll
		\{ \texttt{chocolate} \}   \ll \{ \texttt{chocolate}, \texttt{nachos},\texttt{dips}, \texttt{chillies} \} 
		\ll \ksChoiceAlternatives
	\end{align*}
	One can see that \( \ll \) is \( K \)-minimal and \( \ksChoiceStructure \)-smooth.
	By employing \( \ll \), we obtain the function \( {\triangledown_{\mathrm{Ex}}} = {\triangledown_{\ll}^{K}} \). According to Proposition~\ref{prop:linearchoice_existence}, \( {\triangledown_{\mathrm{Ex}}} \)  is a \( K \)-minimal linear choice function for \( \ksChoiceStructure \). 
    For \( S= \{ \texttt{nachos},\allowbreak \texttt{pretzels},\allowbreak \texttt{dips} \} \) from Example~\ref{ex:snacks1} we obtain \( {\triangledown_{\mathrm{Ex}}} (S) = \{ \texttt{pretzels} \} \); and for \( S' = \{ \texttt{nachos}, \texttt{dips}  \} \) from Example~\ref{ex:snacks3}, we obtain the fallback value \( {\triangledown_{\mathrm{Ex}}} (S') = K \).
    \hfill\( \blacksquare \)
\end{example}

Proposition~\ref{prop:linearchoice_existence} guarantees existence of \( K \)-minimal linear choice function for every restricted choice structure \( \ksChoiceStructure \). 
In the following section, we will axiomatize such choice functions.
 
\section{Axiomatics of \( K \)-minimal Linear Choice Functions}
\label{sec:AxiomaticsKMinimal}

We axiomatise \( K \)-minimal linear choice functions for union-closed and arbitrary restricted choice structures.
Given a restricted choice structure \( \ksChoiceStructure = \tuple{\ksChoiceAlternatives,\ksChoiceDomain,\ksChoiceCoDomain} \), 
we make use of the following postulates for some fixed \( K\in \ksChoiceCoDomain \):
    \begin{center}
        \begin{tabular}{@{}p{0.8cm}p{11.25cm}}

            (\textlabel{SS0}{pstl:SS0a}) & If  \( \ksChoiceCoDomain \cap \ksPowerset{S} \neq \emptyset \) , then \( \triangledown(S) \subseteq S \). \\[0.5em]
            
            (\textlabel{SS1}{pstl:SS1}) & If \( \triangledown(S) \not\subseteq S \), then \( \triangledown(S) = K \). \\[0.5em]
            
            (\textlabel{SS2}{pstl:SS2}) & If \( K \subseteq S \ksTHEN \triangledown(S) = K \). \\[0.5em]
            
            (\textlabel{SS3}{pstl:SS3})  & If \( \triangledown(S_1) \subseteq S_2  \ksAND  \triangledown(S_2)\subseteq S_1 \ksTHEN \triangledown(S_1) = \triangledown(S_2) \). 
            \\[0.5em]

            (\textlabel{SS4}{pstl:SS4})  & If \( \triangledown(S_1) \subseteq S_1 \ksAND S_1 \subseteq S_2 \ksTHEN \triangledown(S_2) \subseteq S_2 \). \\[0.5em]
            
            (\textlabel{SS5}{pstl:SS5}) & If \( \triangledown(S_{i}\cup S_{i+1})  = S_{i} \text{ for \( 0\leq i\leq n \)}\ksTHEN S_0\neq S_{n} \) implies \( \triangledown(S_0\cup S_n)  \neq S_{n} \).\\[0.5em]

            (\textlabel{SS6}{pstl:SS6}) & If \( \triangledown(S_{1}\cup S_{2})  = S_{3} \ksTHEN
            \triangledown(S_{1}\cup S_{3})  = S_{3}  \). 
        \end{tabular}
    \end{center}
By \eqref{pstl:SS0a}, we ensure that a choice is made among the elements of \( S \) whenever \( \ksChoiceCoDomain \) permits this.
The postulate~\eqref{pstl:SS1} describes that either a choice is made among \( S \) or the function falls back to \( K \).
With \eqref{pstl:SS2}, we express that when \( K \) is a subset of \( S \), we must choose exactly \( K \).
The postulate \eqref{pstl:SS3} demands that if \( S_1 \) and \( S_2 \) are mutually supersets of the choices among them, then choosing among \( S_1 \) or \( S_2 \) leads to the same result.
With \eqref{pstl:SS4} we obtain that choosing among elements of \( S \) is inherited to all supersets of \( S \).
With \eqref{pstl:SS5} one prevents cyclic situations among the potential choices.
The postulate \eqref{pstl:SS6} describes that when chooses \( S_3 \) from \( S \), then for each subset \( S' \subseteq S \), we have that \( S_3 \) is chosen from \( S' \cup S_3 \), i.e., the elements of \( S_3 \) are prevalent against the other elements of \( S \).

The first main theorem of this paper is that \eqref{pstl:SS0a}--\eqref{pstl:SS6} exactly characterizes \( K \)-minimal linear choice functions for union-closed restricted choice structures.
Because of the limited space, we present an outline of the proof of the following Theorem~\ref{thm:reptheorem_linearchoice}. 
The full proof is given in the supplemental material.
\begin{theorem}\label{thm:reptheorem_linearchoice}
	Assume the Axiom of Choice.
	Let \( \ksChoiceStructure = \tuple{\ksChoiceAlternatives,\ksChoiceDomain,\ksChoiceCoDomain} \) be a union-closed restricted choice structure and let \( K \in \ksChoiceCoDomain \). 
	A function \( \triangledown : \ksChoiceDomain \to \ksChoiceCoDomain \) is a \( K \)-minimal linear choice function for \( \ksChoiceStructure \) if and only if the axioms \eqref{pstl:SS0a}--\eqref{pstl:SS6} are satisfied\footnote{As usual, satisfaction of \eqref{pstl:SS0a}--\eqref{pstl:SS6} (by \( \triangledown \)) means that the properties described by \eqref{pstl:SS0a}--\eqref{pstl:SS6} hold for all \( S,S_0,S_1,\ldots \in \ksChoiceDomain \) for the specifically considered \(\triangledown\). Note that \( K\) and \(\ksChoiceCoDomain \) in \eqref{pstl:SS0a}--\eqref{pstl:SS6} are externally given, and thus, are not all-quantified.}
\end{theorem}
\begin{proof}[Proof (outline)] {[}\emph{\enquote{\( \Rightarrow \)}}{]} The left-to-right direction amounts to checking the satisfaction of the postulates point-by-point.
	{[}\emph{\enquote{\( \Leftarrow \)}}{]} The right-to-left direction consists of three steps.
	In \emph{Step 1}, one constructs a relation \( {\unlhd} \) on an appropriate subset \( \ksMathcal{E} \subseteq \ksChoiceCoDomain \)  by the following encoding scheme~\cite{KS_Sen1971}:
		\( A_1 \unlhd A_2 \text{ if } \triangledown(A_1\cup A_2) = A_1 \)
	The relation \( {\unlhd} \) is reflexive, antisymmetric, consistent and \( \ksChoiceStructure \)-smooth on \( \ksMathcal{E} \). However, \( \unlhd \) is not a linear order on \( \ksChoiceCoDomain \).
	In \emph{Step 2}, Suzumuras theorem~\cite{KS_Suzumura1976} yields an extension of the relation \( {\unlhd} \) to a \( \ksChoiceStructure \)-smooth linear order \( \lll \) on \( \ksMathcal{E} \). \emph{Step 3} consists of expanding \( \lll \) to a linear order \( \ll \) on \( \ksChoiceCoDomain \)  such that \( \triangledown = \triangledown_{\ll}^{K} \).
\end{proof}

The second main result is an axiomatization of \( K \)-minimal linear choice functions for arbitrary restricted choice structure, by axioms that follow the same ideas as \eqref{pstl:SS1}--\eqref{pstl:SS6}.
\begin{theorem}[Representation Theorem]\label{thm:reptheorem_linearchoiceNonUnion}
	\linebreak[3]Assume the Axiom of Choice.
	Let \( \ksChoiceStructure = \tuple{\ksChoiceAlternatives,\ksChoiceDomain,\ksChoiceCoDomain} \) be a restricted choice structure and let \( K \in \ksChoiceCoDomain \). 
	A function \( \triangledown : \ksChoiceDomain \to \ksChoiceCoDomain \) is a \( K \)-minimal linear choice function for \( \ksChoiceStructure \) if and only if the axioms \eqref{pstl:SS0a}--\eqref{pstl:SS4} and the following axioms are satisfied:		
		\begin{center}
		\begin{tabular}{@{}p{1.1cm}p{10.7cm}}

				(\textlabel{SS5E}{pstl:SS5E}) & If \(  S_{i}\cup S_{i+1} \subseteq S_{i,i+1} \) and \( \triangledown(S_{i,i+1})  = S_{i} \text{ for \( 0\leq i\leq n \)} \), \newline \null\hspace{1cm} then \( S_n\cup S_0 \subseteq S_{n,0} \) and \( S_{n}\neq S_{0} \) imply \( \triangledown(S_{n,0})  \neq S_{n} \). 
				\\[0.5em]

        (\textlabel{SS6E}{pstl:SS6E}) & If \( S_{1,2} = S_{1}\cup S_{2} \) and \( S_{1,3} = S_{1}\cup S_{3} \) and \( \triangledown(S_{1,2})  = S_{3} \), \newline \null\hspace{1cm} then \(
        \triangledown(S_{1,3})  = S_{3}  \). 
				\end{tabular}
		\end{center}
\end{theorem}
The proof of Theorem~\ref{thm:reptheorem_linearchoiceNonUnion} is given in the supplemental material.
The postulates \eqref{pstl:SS5E} and \eqref{pstl:SS6E} are variations of \eqref{pstl:SS5} and \eqref{pstl:SS6} that take into account that the union of elements of \( \ksChoiceDomain \) might not be in  \( \ksChoiceDomain \). 
Note that one demands \(  S_{i}\cup S_{i+1} \subseteq S_{i,i+1} \) in \eqref{pstl:SS5E} and not \(  S_{i}\cup S_{i+1} = S_{i,i+1} \).

\section{Applications in Knowledge Representation}
\label{sec:applyKR}
In this section, we discuss instantiations of union-closed restricted choice structures and the respective interpretation of linear choice functions and Theorem~\ref{thm:reptheorem_linearchoice} in the context of these applications.

\subsection{Linear Choice in Theory Change}
\label{sec:applyTC}
In theory change, one asks how to change a knowledge base \( K \) according to new information \( S \).
Specifically for revision, one wants that \( S \) holds after revising \( K \).
Notably, in the framework by Katsuno and Mendelzon~\cite{KS_KatsunoMendelzon1992}, a change operator is a function \(  \ksLogicLang \times \ksLogicLang \to \ksLogicLang \), where \( \ksLogicLang \) is a language of propositional logic.
We adapt the notion of a change operator to the setting here. When \( \ksChoiceDomain \subseteq \ksPowerset{\ksChoiceAlternatives} \) is a system of subsets of some set \( \ksChoiceAlternatives \), we define a change operator as a function \( \ovee: \ksChoiceDomain \times \ksChoiceDomain \to \ksChoiceDomain \).
The following definition defines a change operator that is based on a choice function.

\begin{definition}\label{def:linearchange}
A change operator \( \ovee: \ksChoiceDomain \times \ksChoiceDomain \to \ksChoiceDomain \) for \( \ksChoiceDomain \) (over \( \ksChoiceAlternatives \)) is called \emph{choice-based} if for each \( K \in \ksChoiceDomain \) there is a choice function \( C_{K} \) for some restricted choice structure \( \ksChoiceStructure_{K} = \tuple{\ksChoiceAlternatives,\ksChoiceDomain,\ksChoiceCoDomain_{K}} \) with \( K \in \ksChoiceCoDomain_{K} \) such that:
\begin{equation*}
    K \ovee S = C_{K}(S)
\end{equation*}
We say \( \ovee \) fits the \( \ksChoiceDomain \)-indexed family of restricted choice structure \( \left\{\tuple{\ksChoiceAlternatives,\ksChoiceDomain,\ksChoiceCoDomain_{K}}\right\}_{K\in\ksChoiceDomain} \).
If each \( C_{K} \) is a \( K \)-minimal linear choice function for \( \ksChoiceStructure \), we say that \( \ovee \) is \emph{linear}.
\end{definition}

Choice functions have already been employed to define change operators~\cite{KS_Rott2001,KS_Haret2020}. 
The novelty here is the restriction on the output side; leading to a model of change operators for agents that are \emph{not able} to conduct certain changes.
In such a setting AGM revision operators are not realizable in general~\cite{KS_SauerwaldThimm2024} or learnable~\cite{KS_BaltagGierasimczukSmets2019}. 
In the following theorem, we employ Theorem~\ref{thm:reptheorem_linearchoice} to give an axiomatization for linear choice-based change operators.

\begin{theorem}\label{thm:linear_change}
	Assume the Axiom of Choice and let \( \ovee: \ksChoiceDomain \times \ksChoiceDomain \to \ksChoiceDomain \) be a change operator for a union-closed set of sets \( \ksChoiceDomain \subseteq \ksPowerset{\ksChoiceAlternatives} \).
	The operator \( {\ovee} \) is linear choice-based if and only if the following postulates are satisfied:	
	\begin{center}
		\begin{tabular}{@{}p{1.2cm}p{10.75cm}}
			(\textlabel{LCR1}{pstl:LCR1}) & \( K \ovee S \subseteq S \) or \( K \ovee S = K \).\\[0.5em]
			
			(\textlabel{LCR2}{pstl:LCR2}) & If \( K \subseteq S \ksTHEN K \ovee S = K \).\\[0.5em]
			
			(\textlabel{LCR3}{pstl:LCR3})  & If \( K \ovee S_{1} \subseteq S_{2}  \ksAND  K \ovee S_{2} \subseteq S_{1} \ksTHEN K \ovee S_{1} = K \ovee S_2 \).\\[0.5em]

			(\textlabel{LCR4}{pstl:LCR4})  & If \( K \ovee S_{1} \subseteq S_1 \ksAND S_1 \subseteq S_2 \ksTHEN K \ovee S_{2} \subseteq S_2 \).\\[0.5em]
			
			(\textlabel{LCR5}{pstl:LCR5}) & If \( K \ovee (S_{i} \cup S_{i+1})  = S_{i} \text{ for \( 0\leq i\leq n \)} \), \newline \null\hspace{2cm} then \( S_0\neq S_{n} \)  implies \( K \ovee (S_0 \cup S_n)  \neq S_{n} \).\\[0.5em]
			
			(\textlabel{LCR6}{pstl:LCR6}) & If \( K \ovee (S_{1} \cup S_{2})  = S_{3} \ksTHEN
			K \ovee (S_{1} \cup S_{3})  = S_{3}  \). 
		\end{tabular}
	\end{center}
\end{theorem}
\begin{proof}[Proof (sketch)]
We show each direction independently.
{[}\emph{\enquote{\( \Rightarrow \)}}{]} Let \( {\ovee} \) be a linear choice-based change operator.
Inspecting Definition~\ref{def:linearchange} reveals that for every \( K \) there is a \( K \)-minimal linear choice function \( C_{K} \) for \( \ksChoiceStructure_{K} \) such that \( 
K \ovee S = C_{K}(S) \).
By employing Theorem~\ref{thm:reptheorem_linearchoice} we obtain that \( C_{K} \) satisfies \eqref{pstl:SS0a}--\eqref{pstl:SS6} for each \( K \). 
From the latter, one obtains easily that \eqref{pstl:LCR1}--\eqref{pstl:LCR6} are satisfied, by substituting \( \triangledown(S) \) by \enquote{\( K \ovee S \)}.
{[}\emph{\enquote{\( \Leftarrow \)}}{]}
Assume that \( \ovee \) satisfies \eqref{pstl:LCR1}--\eqref{pstl:LCR6}. For each fixed \( K \), define a function \( C_{K} : \ksChoiceDomain \to \ksChoiceCoDomain_{K} \) with \( K \ovee S = C_{K}(S) \) and \( \ksChoiceCoDomain_{K} = \bigcup_{S\in\ksChoiceDomain} \{ K \ovee S \} \).
One can see easily that each \( C_{k} \) satisfies \eqref{pstl:SS1}--\eqref{pstl:SS6} by replacing \enquote{\( K \ovee S \)} in every postulate \eqref{pstl:LCR1}--\eqref{pstl:LCR6} by \enquote{\( C_{K}(S) \)}.
The postulate \eqref{pstl:SS0a} is satisfied because \eqref{pstl:SS1} is satisfied and because \( \ksChoiceCoDomain_{K} \) contains only elements that are in the image of \( K \ovee S  \).
By employing Theorem~\ref{thm:reptheorem_linearchoice} we obtain that each \( C_{K} \) is a \( K \)-minimal linear choice function and, hence, that \( \ovee \) is linear.
\end{proof}
The postulates \eqref{pstl:LCR1}--\eqref{pstl:LCR6} are reformulations of the postulates \eqref{pstl:SS1}--\eqref{pstl:SS6}. 
Note that there is no counterpart of \eqref{pstl:SS0a}; inspecting Theorem~\ref{thm:linear_change} reveals that the definition of linear change operator is made in a way that the integrated choice function satisfies already \eqref{pstl:SS0a}. Note that Proposition~\ref{prop:linearchoice_existence} guarantees that linear choice-based change operators always exist.

\begin{corollary}
Assume the Axiom of Choice and let \( \ksChoiceDomain \subseteq \ksPowerset{\ksChoiceAlternatives} \) be a set of sets. 
For any \( \ksChoiceDomain \)-indexed family of restricted choice structure \( \left\{\tuple{\ksChoiceAlternatives,\ksChoiceDomain,\ksChoiceCoDomain_{K}}\right\}_{K\in\ksChoiceDomain} \) with \( K \in \ksChoiceCoDomain_{K} \) for every \( K\in\ksChoiceDomain \) there is a linear choice-based operator that fits that family.
\end{corollary}

We consider the postulates \eqref{pstl:LCR1}--\eqref{pstl:LCR6}.
In theory change the postulates \eqref{pstl:LCR1}---\eqref{pstl:LCR4} are known\footnote{When one reads \( \subseteq \) as some kind of entailment.} as ({Relative Success}), ({Idempotence}), ({Right-Reciprocity}), and ({Successs Monotonicity}).
\eqref{pstl:LCR1} is especially known from non-prioritized revision \cite{KS_Hansson1999b}. 
In non-prioritized revision, one provides change operators for agents that are not willing to accept every new information unquestioned and fully~\cite{KS_FermeHansson1999,KS_HanssonFermeCantwellFalappa2001,KS_KoniecznyPinoPerez2008,KS_SauerwaldBeierle2019a}.
However, the kind of changes we are drafting here differ from non-prioritized change conceptually.
It is not that an agent is not willing to accept some new information, but she cannot because of the restricted outputs.
Technically speaking, in non-prioritized change, the restriction is on how to deal with the inputs within the language.
Restricted choice is about restrictions on the meta-level (especially on the outputs).
Thus, \eqref{pstl:LCR1}  seems to be a postulate that is not exclusive to non-prioritized change; it is more of an expression of dealing with restrictions through a fallback value.
The following example reinterprets our running example in the context of change, demonstrating both cases \eqref{pstl:LCR1} describe.
\begin{example}[Continued from Example~\ref{ex:snacks3}]\label{ex:snacksChange}
    We consider \( K= \{ \texttt{pretzels},\allowbreak \texttt{nachos},\allowbreak \texttt{dips}, \texttt{chillies} \} \) from Example~\ref{ex:snacks3}.
    In the context of change, \( K \) stands for the initial information.
    The linear order \( \ll \) from Example~\ref{ex:snacks3} gives rise to \( \triangledown_{\mathrm{Ex}} \), which we employ for changing \( K \), by setting \( K \ovee S = \triangledown_{\mathrm{Ex}}(S) \). We obtain, e.g., \( K \ovee \{ \texttt{nachos},\allowbreak \texttt{pretzels},\allowbreak \texttt{dips} \} = \{ \texttt{pretzels} \} \) and \( K \ovee \{ \texttt{nachos}, \texttt{dips}  \} = K \).    \hfill\( \blacksquare \)
\end{example}
The postulate \eqref{pstl:LCR2} describes that when the initial information \( K \) is a suitable choice, the operator \( \ovee \) has to output \( K \).
This conforms with the special role the initial beliefs have in many belief change approaches~\cite{KS_FermeHansson2018}.
In a certain way, \eqref{pstl:LCR2} is a basic form of \enquote{minimal change}. 
Note that, minimal change in AGM~\cite{KS_Gaerdenfors1988,KS_AlchourronGaerdenforsMakinson1985} involves that when \( K \cap S \neq \emptyset \), then \( K\ovee S = K \cap S \). However, in the restricted choice setting this is not always possible, as \( K \cap S \notin \ksChoiceCoDomain \) might hold.
The remaining postulates \eqref{pstl:LCR3}--\eqref{pstl:LCR6} deal with the nature of linear orders over sets.
Note that the postulates \eqref{pstl:LCR5} and \eqref{pstl:LCR6} seem to be novel for the change context; they are different from those for linear orders on alternatives~\cite{KS_BelahceneGaigneLagrue2024}.

\subsection{Linear Choice in Argumentation}
\label{sec:applyArg}
\newcommand{\argFramework}{\ensuremath{F}}
\newcommand{\argAttack}{\ensuremath{R}}
\newcommand{\argArguments}{\ensuremath{\ksChoiceAlternatives}}
\newcommand{\argUniverseArguments}{\ensuremath{\ksMathcal{U}}}
\newcommand{\argUniverseAFs}{\ensuremath{\textsf{AF}[\argUniverseArguments]}}
\newcommand{\argSemExtensionRanking}{\ensuremath{\tau}}
\newcommand{\argSemExtension}{\ensuremath{\sigma}}
Formal argumentation \cite{DBLP:journals/aim/AtkinsonBGHPRST17} deals with conflicting pieces of information, where conflicting information is modelled as \emph{arguments} in a discussion.
Dung \cite{DBLP:journals/ai/Dung95} proposed \emph{abstract argumentation frameworks} (AFs) as a directed graph, where arguments are vertices and an \emph{attack} between two arguments is a directed edge.
Formally, given a (possibly infinite) set of arguments \( \argUniverseArguments \), an \emph{abstract argumentation framework} (AF) for \( \argUniverseArguments \) is a directed graph $\argFramework = \tuple{\argArguments,\argAttack} $, where $\argArguments \subseteq \argUniverseArguments$ is a (often finite) non-empty set of arguments and $\argAttack$ is an \emph{attack relation} $\argAttack \subseteq \argArguments \times \argArguments$. With $\argUniverseAFs$ we denote the set of all abstract argumentation frameworks for \( \argUniverseArguments \).

There are different approaches to the semantics of abstract argumentation frameworks~\cite{baroni2018handbook}. 
Most prominently, extension-based semantics assign to each AF \( \argFramework = \tuple{\argArguments,\argAttack} \) a set of sets of arguments \( \argSemExtension(\argFramework) \subseteq \ksPowerset{\argArguments} \) (called the \( \argSemExtension \)-extensions of \( \argFramework \)).
The intended meaning is that \( \argSemExtension(\argFramework) \) represents the viable solutions of the conflict resolution between the arguments in \( \argFramework \).
A generalization of extension-based semantics are extension-ranking semantics, which additionally equip the extensions with an ordering~\cite{KS_SkibaRienstraThimmHeyninckKernIsberner2021}, i.e., functions \( \ensuremath{\tau} \) that map each AF to an ordering \( \leq^\ensuremath{\tau}_{\argFramework} \) over some subsets of \( \ksPowerset{\argArguments} \).
Extension-ranking semantics provide much more structure to the solutions given by extension-based semantics, i.e., that some extensions are \enquote{better} than others.
We define a semantics for AFs that is based on choice functions.
\begin{definition}
    A \emph{choice-based extension semantics} for \( \argUniverseArguments \) is a function $\Pi$ that maps each AF $\argFramework = \tuple{\argArguments,\argAttack}$ for \( \argUniverseArguments \) to a choice function $\Pi_{\argFramework} : \ksMathcal{P}(\argArguments) \to \ksChoiceCoDomain_{\argFramework} $ for some choice structure $\ksChoiceStructure_{\argFramework} = \tuple{\argArguments, \ksMathcal{P}(\argArguments), \ksChoiceCoDomain_{\argFramework}}$ where \( \ksChoiceCoDomain_{\argFramework} \) is non-empty. 
    We say that \( \Pi \) is \emph{linear}, if every \( \Pi_{\argFramework} \) is a \( K_{\argFramework} \)-minimal linear choice function for some \( K_{\argFramework} \in \ksChoiceCoDomain_{\argFramework} \).
\end{definition}

\begin{figure}[t]
    \begin{center}
        \begin{tikzpicture}[thick]
            \node (nachos) at (6,0) {\texttt{nachos}};
            \node (dips) at (3,0) {\texttt{dips}};
            \node (chillies) at (0,0) {\texttt{chillies}};
            \node (chocolate) at (1.5,-1.5) {\texttt{chocolate}};
            \node (pretzels) at (4.5,0.-1.5) {\texttt{pretzels}};
            
            \draw (pretzels) edge [-{latex}] (nachos);
            \draw (pretzels) edge [-{latex}, bend right=10] (dips);
            \draw (dips) edge [-{latex}, bend right=10] (pretzels);
            \draw (dips) edge [-{latex}] (chocolate);
            \draw (chocolate) edge [-{latex}] (chillies);
        \end{tikzpicture}
    \end{center}
    \caption{Illustration of the abstract argumentation framework from Example~\ref{ex:af}.}
    \label{fig:ex_af}
\end{figure}
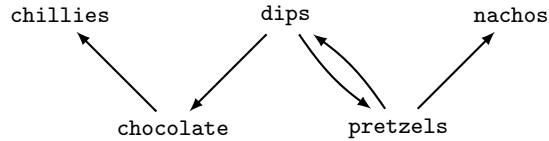
Choice-based extension semantics provide for every argumentation framework \( \argFramework \) a restricted choice function $ \Pi_{\argFramework}$ over the arguments of \( \argFramework \).
An instantiation for \( \ksChoiceCoDomain_{\argFramework} \) could be the extensions of \( \argFramework \) (with respect to some extension-based semantics).
Given that, one can interpret a choice $ \Pi_{\argFramework}(E)$ for some \( E \subseteq \argArguments \) as an answer to the question of \enquote{which arguments in \( E \) (if any) are conflict free?}.
Some AFs can have a big number of arguments, hence it is of interest to only look at a selection of extensions that only concern these arguments. 
This is also known extension selection~\cite{DBLP:conf/ecsqaru/KoniecznyMV15,DBLP:conf/kr/BonzonDKM18,KS_SkibaRienstraThimmHeyninckKernIsberner2021}.

Next, we consider the axiomatic linear choice-based extension semantics provided by Theorem~\ref{thm:reptheorem_linearchoice}.
\begin{theorem}    
    Assume the Axiom of Choice and let $\Pi$ be a choice-based extension semantics for \( \argUniverseArguments \).
    The semantics \( {\Pi} \) is linear if and only if the following postulates are satisfied (for a suitable \( K_{\argFramework} \) for each \( \argFramework \in \argUniverseAFs \)):	%
    \begin{center}%
	\begin{tabular}{@{}p{1.2cm}p{10.75cm}@{}}		

		(\textlabel{LCA1}{pstl:LCA1}) & \( \Pi_{\argFramework}(E) \subseteq E \) or \( \Pi_{\argFramework}(E) =  K_{\argFramework} \).\\[0.5em]
		
		(\textlabel{LCA2}{pstl:LCA2}) & If \( K_{\argFramework} \subseteq  \Pi_{\argFramework}(E) \ksTHEN  \Pi_{\argFramework}(E) = K_{\argFramework} \).\\[0.5em]
		
		(\textlabel{LCA3}{pstl:LCA3})  & If \(  \Pi_{\argFramework}(E_1) \subseteq E_2 \ksAND  \Pi_{\argFramework}(E_2) \subseteq E_1 \ksTHEN  \Pi_{\argFramework}(E_1) =  \Pi_{\argFramework}(E_2)  \).\\[0.5em]

		(\textlabel{LCA4}{pstl:LCA4})  & If \(  \Pi_{\argFramework}(E_1) \subseteq E_1    \ksAND E_1 \subseteq E_2 \ksTHEN  \Pi_{\argFramework}(E_2) \subseteq E_2 \).\\[0.5em]
		
		(\textlabel{LCA5}{pstl:LCA5}) & If \(  \Pi_{\argFramework}(E_i \cup E_{i+1}) = E_i \text{ for \( 0\leq i\leq n \)} \), \newline \null\hspace{1cm} then \( E_0\neq E_{n} \) implies \(  \Pi_{\argFramework}(E_0 \cup E_n) \neq  E_{n} \).\\[0.5em]

		(\textlabel{LCA6}{pstl:LCA6}) & If \(  \Pi_{\argFramework}(E_1 \cup E_2) = E_3 \ksTHEN
		 \Pi_{\argFramework}(E_1 \cup E_3) = E_3  \).
	\end{tabular}
\end{center}
\end{theorem}
\begin{proof}
	We show each direction independently.
    {[}\emph{\enquote{\( \Rightarrow \)}.}{]} 
    Let \(\Pi\) be a linear choice-based extension semantics. Then for every  \(\argFramework  \in  \argUniverseAFs \) there is a  \( K_{\argFramework} \)-minimal linear choice function \(\Pi_{\argFramework}\) for \(R_{\argFramework}\). By employing Theorem~\ref{thm:reptheorem_linearchoice}, we obtain that \(\Pi_{\argFramework}\) satisfies  \eqref{pstl:SS1}--\eqref{pstl:SS6} for each \( K_{\argFramework} \), which yields the satisfaction of \eqref{pstl:LCA1}--\eqref{pstl:LCA6}.
    {[}\emph{\enquote{\( \Leftarrow \)}.}{]} Assume that \(\Pi\) satisfies  \eqref{pstl:LCA1}--\eqref{pstl:LCA6}. 
    This implies that every function \( \Pi_{\argFramework} \) satisfies \eqref{pstl:SS1}--\eqref{pstl:SS6}.
    Now define \( \ksChoiceCoDomain_{\argFramework} = \bigcup_{E \in \ksMathcal{P}(\argArguments)}\{\Pi_{\argFramework}(E)\}\). 
    Then, the satisfaction of \eqref{pstl:SS0a}  by  \( \Pi_{\argFramework} \) is guaranteed by satisfaction of \eqref{pstl:SS1} and by the definition of  \( \ksChoiceCoDomain_{\argFramework} \).
    Because \(\Pi_{\argFramework}\) satisfies \eqref{pstl:SS0a}--\eqref{pstl:SS6}, we obtain that \(\Pi_{\argFramework}\) is a \(K_{\argFramework}\)-minimal linear choice function from Theorem~\ref{thm:reptheorem_linearchoice}, and thus, that $\Pi$ is linear. 
\end{proof}

A direct connection to principles for abstract argumentation semantics~\cite{KS_BaroniGiacomin2007,KS_TorreVesic2017} is not obvious.
This is because, from the perspective of argumentation, much of the quality of the results \( \Pi_{\argFramework}(E) \) depends on the internally chosen set \( \ksChoiceCoDomain_{\argFramework} \).
However, for the discussion here, we just assume that each element in \( \ksChoiceCoDomain_{\argFramework} \) is well-behaving, e.g., conflict-free, i.e., for \( E \in \ksChoiceCoDomain_{\argFramework} \) we have \( a,b\in E \) implies that \( (a,b) \notin \argAttack \)~\cite{DBLP:journals/ai/Dung95}. Hence, under this assumption, our restriction of \( \Pi_{\argFramework}(E) \) to \( \ksChoiceCoDomain_{\argFramework} \) guarantees well-behaving outputs.

\begin{example}[Continued from Example~\ref{ex:snacksChange}]\label{ex:af}
    We interpret Examples~\ref{ex:snacks1}--\ref{ex:snacks3} in the context of argumentation. Here, the elements of \( \ksChoiceAlternatives \) are understood as arguments and the elements of \( \ksChoiceCoDomain \) are meant to be the \enquote{feasible} subset of arguments. The linear order \( \ll \) provides an ordering on \( \ksChoiceCoDomain \), where the smaller elements are more \enquote{feasible}.
    An argumentation framework that suits this scenario is the framework \( \argFramework = \tuple{A,\argAttack} \) with \( \argAttack = \{ (\texttt{pretzels},\texttt{nachos}), (\texttt{pretzels},\texttt{dips}), (\texttt{dips},\allowbreak\texttt{pretzels}),
    \allowbreak(\texttt{dips},\texttt{chocolate}), (\texttt{chocolate},\texttt{chillies}) \} \).
    For an illustration, see Figure~\ref{fig:ex_af}.
Example of  outputs of the choice-based extension semantics \( \Pi \) are \( \Pi_{\argFramework}(\allowbreak\{ \texttt{nachos},\allowbreak \texttt{pretzels},\allowbreak \texttt{dips} \}) \allowbreak=\allowbreak \{ \texttt{pretzels} \} \) and \( \Pi_{\argFramework}(\{ \texttt{nachos}, \texttt{dips}  \}) = K \). 
   \hfill\( \blacksquare \)
\end{example}
Choice-based extension semantics seem to be an interesting novel type of abstract argumentation semantics. 
However, how linear choice-based extension semantics handles fallback values does not seem appropriate. 
In Example~\ref{ex:af}, we observe the behaviour of \( \Pi_{\argFramework}(\{ \texttt{nachos}, \texttt{dips}  \}) = K \) in Example~\ref{ex:af}.
This is because the fallback value is the minimal element of the respective linear order.
Thus, one might not want the fallback value to be the minimal element in abstract argumentation.
For instance, for some semantics, the empty extension \( \emptyset \) would be a suitable fallback value. However, when \( K_{\argFramework}=\emptyset \) is the minimal element, then \( \Pi_{\argFramework}(E) = \emptyset  \) holds for all \( E \in \ksChoiceDomain \).

\section{Conclusion}
\label{sec:conclusion}

In this paper, we considered the novel setting of restricted choices.
A full axiomatization (Theorem~\ref{thm:reptheorem_linearchoiceNonUnion}) is given for \( K \)-minimal linear choice functions, i.e., those choice functions that can be represented by a linear order over sets of sets where a fallback value is encoded as the minimal element.  As given by Proposition~\ref{prop:linearchoice_existence}, such choice functions always exist regardless of which restrictions on choices are given.
We showed that Theorem~\ref{thm:reptheorem_linearchoiceNonUnion} can be simplified to Theorem~\ref{thm:reptheorem_linearchoice}  for choice structures where the input is closed under union, i.e., in this setting, \eqref{pstl:SS1}--\eqref{pstl:SS6} fully characterise \( K \)-minimal linear choice functions. 
We give a first hint on the applications of linear choice functions in theory change and argumentation. 

In future work, we will transfer the approach of this paper to the area of non-monotonic reasoning, especially preferential reasoning~\cite{KS_KrausLehmannMagidor1990}.
It turns out that transferring preferential reasoning to other domains is axiomatically challenging~\cite{KS_SauerwaldKontinen2024}.
As non-monotonic reasoning and belief change are known to be connected~\cite{KS_GaerdenforsMakinson1994}, we are optimistic that our approach here might have applications in the axiomatization of non-monotonic reasoning approaches.
Furthermore, for the area of argumentation, the idea of choice-based semantics sketched here is a promising and interesting area to explore.

Another avenue of future work is further exploration of the framework itself. We will consider alternative fallback behaviours for choice functions and explore their conceptual relevance. For that, identifying natural fallback behaviours for different applications in knowledge representation and beyond will be very helpful.

\subsubsection*{\ackname}
We thank the reviewers of this paper for their valuable and constructive feedback, which significantly helped us to improve the paper.
The research reported here is partially funded by the Deutsche Forschungsgemeinschaft (DFG, German Research Foundation)---project 506604007 (Algorithms for Inconsistency Measurement) and project 465447331 (Explainable Belief Merging).  Kenneth Skibka was supported by project 506604007 and Kai Sauerwald was supported by project 465447331.
Eduardo Fermé was partially supported by UID/04516/NOVA Laboratory for Computer Science and Informatics (NOVA LINCS) with the financial support of FCT.IP. This work is based on the research supported in part by the National Research Foundation of South Africa (REFERENCE NO: SAI240823262612).
\bibliographystyle{splncs04}
\bibliography{bibexport}

\subsubsection*{Preprint Disclaimer}
This preprint has not undergone any post-submission improvements or corrections. The Version of Record of this contribution is~\cite{KS_SauerwaldSkibaFermeMeyer2025}, published in Logics in 
Artificial Intelligence (JELIA 2025), LNCS 16094, and is available online at \url{https://doi.org/10.1007/978-3-032-04590-4_5}

\clearpage
\appendix

\section{Supplementary Material}
In this appendix, we consider the full proofs for
 Theorem~\ref{thm:reptheorem_linearchoice} and  Theorem~\ref{thm:reptheorem_linearchoiceNonUnion}.
First, in Section~\ref{sec:background_extensions}, we present the background and notions we will use.
In the remaining parts, we present the proof for Theorem~\ref{thm:reptheorem_linearchoiceNonUnion}.
Proposition~\ref{prop:rightdirection} in Section~\ref{sec:appproof} provides one direction of Theorem~\ref{thm:reptheorem_linearchoiceNonUnion}, by showing that linear choice functions satisfy the postulates \eqref{pstl:SS1}--\eqref{pstl:SS4}, \eqref{pstl:SS5E} and \eqref{pstl:SS6E}.
Section~\ref{sec:appproof2} will provide a series of Propositions that will ultimately lead to Theorem~\ref{thm:reptheorem_linearchoiceNonUnion}.
Theorem~\ref{thm:reptheorem_linearchoice} will be then derived from Theorem~\ref{thm:reptheorem_linearchoiceNonUnion}.

\subsection{Background on Relations and Order-Extensions}\label{sec:background_extensions}
The transitive closure \( \mathsf{tc}({\preceq}) \) of a relation \( {\preceq}  \) on \( X \) is defined by \( \mathsf{tc}({\preceq})=\bigcup_{i\in\mathbb{N}} {\preceq}^i \), whereby, for \( n\in\mathbb{N}^{+} \),
\begin{align*}
    {\preceq}^0& ={\preceq} & &\ksAND & {\preceq}^n & = {\preceq}^{n-1} \cup \{\ (x_1,x_3) \in X \times X  \mid x_1 \preceq^{n-1} x_2,\ x_2 \preceq^{n-1} x_3 \ \} \ .
\end{align*}
For every relation \( {\preceq} \), the transitive closure of \( {\preceq} \) is the smallest superset of \( {\preceq} \) that is \hyperref[pstl:transitive]{transitive}.
We will make use of the following theorem, stating that a relation \( \preceq \) is consistent if and only if \( \preceq \) can be extended to a total preorder, while retaining the strict part of~\( \preceq \).
For that, we have to assume the \emph{Axiom of Choice}.
Note that the Axiom of Choice is a common assumption in computer science.
\begin{theorem}[Suzumura's theorem~\cite{KS_Suzumura1976}]\label{thm:Suzumura_extension}
Assume the Axiom of Choice.
	A relation \( \preceq \) on a set \( X \) is consistent if and only if there exist a total preorder \( {\preceq^\mathrm{tpo}} \) on \( X \) such that
	\begin{itemize}
		\item if \( x \preceq y  \), then \( x \preceq^\mathrm{tpo} y \), and
		\item if \( x \prec y  \), then \( x \prec^\mathrm{tpo} y  \).
	\end{itemize}
\end{theorem}
\noindent 

In this paper we will also derive linear orders from total preorders such that the strict part is retained.
For that we will assume that the \emph{Linear Ordering Principle} is satisfied, i.e., for every set there exists a linear order on its elements.
\begin{proposition}\label{prop:linear_extension_finite}
    Assume that the Linear Ordering Principle holds.
	For every total preorder \( \preceq \) on a set \( X \), with strict part \( \prec \), there exists a linear order \( {\ll} \) on \( X \) such that
\( x \prec y  \) implies \( x \ll y  \) for all \( x,y\in X \).
\end{proposition}
\begin{proof}
	Because the linear ordering principle holds, there exist a linear order \( {\sqsubseteq} \subseteq X\times X \) on \( X \).
	We define \( {\ll} \subseteq X\times X \) by employing \( {\subseteq} \) as follows:
	\begin{equation*}
		x \ll y \text{ if } x  \preceq y \text{ and } (\ y \preceq x \text{ implies } x \sqsubseteq y \ )
	\end{equation*}
	Clearly, \( x \prec y  \) implies \( x \ll y  \) for all \( x,y\in X \).
\end{proof}
Note that the \emph{Axiom of Choice} implies the Linear Ordering Principle.
Moreover, when one only considers finite sets \( X \), Proposition~\ref{prop:linear_extension_finite} holds also when ones do not assume the Linear Ordering Principle.

Finally, we use additional notion to improve the readability of the proofs.
When \( \ll \) is a linear order on \( \ksMathcal{E} \), then \( \minOf{S}{\ll} \) will either be a singleton set or empty.
For such linear orders, we use \( \mins(\ll) \) as shorthand for \( \minsOf{A}{\ll} \), whereby we define \( \minsOf{S}{\ll} \) as the single element in \( \minOf{S}{\ll} \) if is a singleton set; otherwise, we say that \( \minsOf{S}{\ll} \) is undefined, i.e.:
\begin{equation*}
    \minsOf{S}{\ll} = \begin{cases}
	        E & \ksIF \minOf{S}{\ll} = \{ E \}\\
	        \text{undefined} & \ksOtherwise
	    \end{cases}
\end{equation*}
By doing so, we can easily address with \( \minsOf{S}{\ll} = E \) the one and only single element in \( \minOf{S}{\ll}=\{ E \} \), i.e., we avoid the explicit unboxing of \( E \) from \( \{ E \} \).
Note that \( \minsOf{S}{\ll} \) is also undefined if \( \minOf{S}{\ll} \) contains more than one element, which can appear when \( \ll \) is not a linear order.

\subsection{\( K \)-Minimal Linear Choice Functions to Postulates}\label{sec:appproof}

We start with the left-to-right direction of Theorem~\ref{thm:reptheorem_linearchoiceNonUnion}.

 \begin{proposition}\label{prop:rightdirection}
	Let \( \ksChoiceStructure = \tuple{\ksChoiceAlternatives,\ksChoiceDomain,\ksChoiceCoDomain} \) be a restricted choice structure. 
	Every \( K \)-minimal linear choice function \( \triangledown : \ksChoiceDomain \to \ksChoiceCoDomain \) for \( \ksChoiceStructure \) satisfies \eqref{pstl:SS0a}--\eqref{pstl:SS4}, \eqref{pstl:SS5E} and \eqref{pstl:SS6E}.
 \end{proposition}
\begin{proof}
	Let  \( \triangledown : \ksChoiceDomain \to \ksChoiceCoDomain \) be a linear choice function for \( \ksChoiceStructure \).
	This means that there is a linear order \( {\ll} \)  on \( \ksChoiceCoDomain \) that is \( \ksChoiceStructure \)-smooth such that \( {\triangledown} = {\triangledown_{\ll}} \).
	We show satisfaction of the postulates:
	\begin{description}
		\item[\normalfont{[}\eqref{pstl:SS0a} is satisfied{]}] 
		We have to show that if \( \ksChoiceCoDomain \cap \ksPowerset{S} \neq \emptyset \), then \( \triangledown(S) \subseteq S \) holds. 
		Clearly, \( \ksChoiceCoDomain \cap \ksPowerset{S} \neq \emptyset \) implies that there is some \( E \in \ksChoiceCoDomain \) with \( E \subseteq S \). Thus, we also have that \( \minsOf{S}{\ll} \) is non-empty. Inspecting Definition~\ref{def:linear_choice_function} reveals that this means that \( \triangledown(S) \subseteq S \).

		\item[\normalfont{[}\eqref{pstl:SS1} is satisfied{]}] 
		We have to show that \( \triangledown(S) \subseteq S \) or \( \triangledown(S) = K \) holds. There are two cases.
		The first case is \( K \subseteq S \). 
		In this case we have that \( K \) is the globally minimal element of \( \ll \).
		Hence, we obtain that \( \triangledown(S) = K = \triangledown_{\ll}(S) = \minsOf{S}{\ll} \).
		If \( K \not\subseteq S \) holds, then there are two subcases. 
		The first subcase is that there no \( E \in \ksChoiceCoDomain \) such that \( E \subseteq S \).
		In this case we have \( \minOf{S}{\ll} = \emptyset \), and thus, by definition of \( \triangledown_{\ll} \), we obtain \( \triangledown_{\ll}(S)=\triangledown(S)= K  \).
		The second subcase is that there an \( E \in \ksChoiceCoDomain \) such that \( E \subseteq S \).
		In this case we have \( \minsOf{S}{\ll} \subseteq S \), because \( \ll \) is \( N \)-smooth.
		Consequently, we obtain \( \triangledown(S) \subseteq S \) because of  \( {\triangledown} = {\triangledown_{\ll}} \).
		\item[\normalfont{[}\eqref{pstl:SS2} is satisfied{]}] 
		We have to show that \( K\subseteq S \) implies \( \triangledown(S)=K \).
		Note that \( \mins(\ll)=K \) holds, because \( \triangledown \) is a linear choice function for \( N \).
		Consequently, for each \( K\subseteq S \) we obtain \( \triangledown(S)=\minsOf{S}{\ll}=K  \).

		\item[\normalfont{[}\eqref{pstl:SS3} is satisfied{]}] 
		We have to show that \(  \triangledown(S_{1}) \subseteq S_{2} \ksAND  \triangledown(S_{2})\subseteq S_{1} \) together imply \( \triangledown(S_{1}) = \triangledown(S_{2}) \).
		Assume that \(  \triangledown(S_{1}) \subseteq S_{2} \ksAND  \triangledown(S_{2})\subseteq S_{1} \) hold.
		Because \( \triangledown \) satisfies \eqref{pstl:SS1}, we have  \( \triangledown(S) \subseteq S \) or \( \triangledown(S) = K \) for each \( S\in\ksChoiceCoDomain \).
		
		The case of \( \triangledown(S_{1}) = K \) or \( \triangledown(S_{2}) = K \). We will consider only the case of \( \triangledown(S_{1}) = K \), the proof for the case of \( \triangledown(S_{2}) = K \) is analogue.
		From \( \triangledown(S_{1}) \subseteq S_{2} \), we obtain that \( K \subseteq S_{2} \) holds. 
		Because \( \triangledown \) satisfies \eqref{pstl:SS2}, we obtain \( \triangledown(S_{2}) = K  \) from \( K \subseteq S_{2} \).
		Hence, we have \( \triangledown(S_{1}) = \triangledown(S_{2}) \).
		
		The case of \( \triangledown(S_{1}) \subseteq S_{1} \) and  \( \triangledown(S_{2}) \subseteq S_{2} \).  
		Let \( M = {\minsOf{S_{1}}{\ll}} \) and let \( N = {\minsOf{S_{2}}{\ll}} \). 
		Clearly, we have that \( M = \triangledown(S_{1}) \) and \( N = \triangledown(S_{2}) \) hold. 
		If \( M = N \), then we are done. We continue for the case of \( M \neq N \).
		From  our assumptions, we obtain that \( M \subseteq S_{2}  \) and that \( N \subseteq S_{1}  \) holds. 
		Now, recall that every linear order is total and antisymmetric, and thus, also \( \ll \).
		Hence, we have that exactly one of \( M \ll N \) or \( N \ll M \) holds.
		This yields a contradiction, because \( M \ll N \) contradicts \( N = \minsOf{S_{2}}{\ll} \), and \( N \ll M \) contradicts \( M = \minsOf{S_{1}}{\ll} \).

		\item[\normalfont{[}\eqref{pstl:SS4} is satisfied{]}] 
		We have to show that \( \triangledown(S_{1}) \subseteq S_{1} \) and \( S_{1} \subseteq S_{2} \) implies \( \triangledown(S_{2}) \subseteq S_{2} \).
		Assume that \( \triangledown(S_{1}) \subseteq S_{1} \) and \( S_{1} \subseteq S_{2} \) hold.
		First, we show that \( \minsOf{S_{1}}{\ll} \) is defined.
		We consider the two cases of \( \triangledown(S_{1}) \neq K \) and \( \triangledown(S_{1}) = K \).
		From Equation~\eqref{eq:triangledownLLdefinition} we obtain that \( \triangledown(S_{1}) \subseteq S_{1} \) is only possible in the case of \( \triangledown(S_{1}) \neq K \), if \( \minsOf{S_{1}}{\ll} \) is defined. For the case of \( \triangledown(S_{1}) = K \), recall that \( \min({\ll})=K \) holds because \( \ll \) is \( \ksChoiceStructure \)-smooth. Consequently, we obtain that  \( \minsOf{S_{1}}{\ll} \) is defined from \( \triangledown(S_{1}) = K \subseteq S_{1} \).        
		Now, because \( \minsOf{S_{1}}{\ll} \) is defined, we obtain that \( \minsOf{S_{2}}{\ll} \) is defined from \( S_{1}\subseteq S_{2} \) and Equation~\eqref{eq:triangledownLLdefinition}. 
		Consequently, 
		we obtain that \( \triangledown(S_{2}) \subseteq S_{2} \) holds.

        \item[\normalfont{[}\eqref{pstl:SS5E} is satisfied{]}] 
		We have to show that for
		\begin{align*}
			S_{0} \cup S_{1} & \subseteq S_{0,1}\\
			 & \vdots \\
			S_{n-1} \cup S_{n} & \subseteq S_{n-1,n}\\
			S_{n} \cup S_{0} & \subseteq S_{n,0}
		\end{align*}
		with		
		 \( \triangledown(S_{0,1}) = S_{0}, \ldots, \triangledown(S_{n-1,n}) = S_{n-1} \) and \( S_0 \neq S_{n} \) together imply \( \triangledown(S_{n,0})  \neq S_{n} \).
		The proof is by contradiction. Henceforth, towards a contradiction, assume that \( \triangledown(S_{n,0}) = S_{n} \) holds, as well as 
		\begin{equation*}
			\triangledown(S_{0,1}) = S_{0},\ \ldots,\ \triangledown(S_{n-1,n}) = S_{n-1}
		\end{equation*}		
		and \( S_0 \neq S_{n} \).
		
		First, we show that \( \minsOf{S_{i,i+1 \,\mathrel{\mathrm{mod}}\, (n+1)}}{\ll} \) is defined for all \( i \in \{ 0, \ldots, n \} \). Towards a contradiction, assume that \( \minsOf{S_{i,i+1 \,\mathrel{\mathrm{mod}}\, (n+1)}}{\ll} \) is undefined.
		Clearly, by considering Equation~\eqref{eq:triangledownLLdefinition} we obtain that \( {\triangledown(S_{i,i+1 \,\mathrel{\mathrm{mod}}\, (n+1)}) = K} \).
		The last observation together with \( \triangledown(S_{i,i+1 \,\mathrel{\mathrm{mod}}\, (n+1)})= S_{i} \) implies that \( {S_{i} = K} \) holds.
		And consequently, because \( \ll \) is a \( K \)-minimal \( \ksChoiceStructure \)-smooth linear order, we have \( {\min(\ll)} = \{ K \} \).
		We obtain that \( {\minsOf{S_{i,i+1 \,\mathrel{\mathrm{mod}}\, (n+1)}}{\ll} = K} \) is defined.
		
		For what remains of proving that \eqref{pstl:SS5E} is satisfied, we assume that \( S_{i} \neq S_{j} \) holds for all \( i,j\in\{0,\ldots, n\} \). 
		This is valid, because we have \( S_{0} \neq S_{n} \) and whenever \( S_{i} = S_{j} \) holds with \( i < j \), we can just consider this proof just for the sets \(  S_{1}, \ldots, S_{i}, S_{j+1},\ldots S_{n}\).
		
		Because, \( \minsOf{S_{i,i+1 \,\mathrel{\mathrm{mod}}\, (n+1)}}{\ll}  \) is defined, we have \[ \triangledown(S_{i,i+1 \,\mathrel{\mathrm{mod}}\, (n+1)}) = \minsOf{ S_{i,i+1 \,\mathrel{\mathrm{mod}}\, (n+1)} }{\ll} \ . \] 
		Consequently, because we have \( S_{i} \neq S_{i+1 \,\mathrel{\mathrm{mod}}\, (n+1)} \), we obtain \( S_{j} \ll S_{j+1} \) for all \( j\in\{0,\ldots,n-1\} \) from \( \triangledown(S_{j,j+1}) = S_{j} \).
		Moreover, \( \triangledown(S_{n,0}) = S_{n}  \) yields \( S_{n} \ll S_{0} \).
		Which shows that we have \( S_{0} \ll S_{1} \ll \ldots \ll S_{n} \ll S_{0} \) and thus, that \( \ll \) violates the \hyperref[pstl:Suzumura_consistency]{consistency} condition. We obtain a contradiction, because \( \ll \) is a \hyperref[pstl:transitive]{transitive} relation, yet not \hyperref[pstl:Suzumura_consistency]{consistent}, which is impossible.
        \item[\normalfont{[}\eqref{pstl:SS6E} is satisfied{]}] 
        First, note that \( S_{1,2}=S_{1} \cup S_{2} \) and \( S_{1,3}=S_{1} \cup S_{3} \) ensure existence of the corresponding unions.
        We have to show that \( \ksChoiceFunction(S_{1} \cup S_{2}) = S_{3} \) implies \(
        \ksChoiceFunction(S_{1} \cup S_{3}) = S_{3} \).
        Assume that \( \ksChoiceFunction(S_{1} \cup S_{2}) = S_{3} \) holds.
        Because \( \ksChoiceFunction \) satisfies \eqref{pstl:SS1}, we have \( \ksChoiceFunction(S_{1} \cup S_{2}) = K \) or \( \ksChoiceFunction(S_{1} \cup S_{2}) \subseteq S_{1} \cup S_{2} \).
        
        We start with the case that \( \triangledown(S_{1} \cup S_{2}) = K \) holds.
        Observe that in this case, we have \( K = S_{3} \), and thus we have \( K \subseteq S_{1} \cup S_{3} \).
        Consequently, we obtain \(   \triangledown(S_{1} \cup S_{3}) = K = S_{3} \) from \eqref{pstl:SS2}.
        
        We continue with the case of \( \ksChoiceFunction(S_{1} \cup S_{2}) \subseteq S_{1} \cup S_{2} \) and \( \ksChoiceFunction(S_{1} \cup S_{2}) \neq K \).
        This case is only possible if \( \minsOf{S_{1} \cup S_{2}}{\ll}  \) is defined.
        Consequently, from Equation~\ref{eq:triangledownLLdefinition} and \( \ksChoiceFunction(S_{1} \cup S_{2}) = S_{3} \), we obtain \( \minsOf{S_{1} \cup S_{2}}{\ll} = S_{3} \). Because of the latter, also \( \minsOf{S_{1} \cup S_{3}}{\ll}  \) is defined and thus,  \( \ksChoiceFunction(S_{1} \cup S_{3}) = \minsOf{S_{1} \cup S_{3}}{\ll} \).
        Now, towards a contradiction to \( \ksChoiceFunction(S_{1} \cup S_{3}) =  S_{3} \) assume \( \ksChoiceFunction(S_{1} \cup S_{3}) \neq  S_{3} \).
        Clearly, this means there is some \( S \) with \( S_{3} \neq S \) and  \( {S} = \ksChoiceFunction(S_{1} \cup S_{3}) = {\minsOf{S_{1} \cup S_{3}}{\ll} \neq  S_{3}}  \).
        Because of this, we obtain \( {\minsOf{S_{1} \cup S_{3}}{\ll}} = S \subseteq S_{3} \) from Equation~\eqref{eq:triangledownLLdefinition}.
        Observe also that we have \( S \ll S_{3} \), as otherwise, we wouldn't have \( S = {\minsOf{S_{1} \cup S_{3}}{\ll}} \).
        Because we have \( \minsOf{S_{1} \cup S_{2}}{\ll} = S_{3} \) and \( \ksChoiceFunction(S_{1} \cup S_{2}) \neq K \), we obtain \( S_{3} \subseteq S_{1} \cup S_{2} \) from \eqref{pstl:SS1}.
        Furthermore, observe that \( S_{1} \cup S_{3} \subseteq S_{1} \cup S_{2} \) holds, because \( S_{3} \subseteq S_{1} \cup S_{2} \) holds. 
        Consequently, we also have that \( S \subseteq S_{1} \cup S_{2} \).
        From the chain of though above obtain that \( S \in \minOf{S_{1} \cup S_{2}}{\ll} \) and  \( S_{3} \in \minOf{S_{1} \cup S_{2}}{\ll} \) holds a that the same time, which is impossible, because \( \ll \) is a linear order. \qedhere
	\end{description}
\end{proof}

From Proposition~\ref{prop:rightdirection}, we obtain the following corollary for union-closed choice structures.
\begin{corollary}\label{prop:rightdirection_unionclosed}
	Let \( \ksChoiceStructure = \tuple{\ksChoiceAlternatives,\ksChoiceDomain,\ksChoiceCoDomain} \) be a union-closed restricted choice structure. 
	Every \( K \)-minimal linear choice function \( \triangledown : \ksChoiceDomain \to \ksChoiceCoDomain \) for \( \ksChoiceStructure \) with fallback \( K \) satisfies \eqref{pstl:SS0a}--\eqref{pstl:SS6}.
\end{corollary}

\subsection{Postulates to \( K \)-Minimal Linear Choice Functions}\label{sec:appproof2}
In the following, we provide an approach to obtain a linear order \( \ll \) that encodes a choice function \( \ksChoiceFunction \) that satisfies \eqref{pstl:SS1}--\eqref{pstl:SS4}, \eqref{pstl:SS5E} and \eqref{pstl:SS6E}.
This order \( \ll \) will be obtained by a multistep process, consisting of a constructive part and a non-constructive part.
In the constructive part we construct a smooth relation that encodes the behaviour of the choice function and is \enquote{nearly} a linear order.
We call such relations \( \ksChoiceStructure \)-compatible.

\newpage
\begin{definition}\label{def:BUCcompatible}
	Let \( \ksChoiceStructure = \tuple{\ksChoiceAlternatives,\ksChoiceDomain,\ksChoiceCoDomain} \) be a restricted choice structure and let be a function \( \ksChoiceFunction : \ksChoiceDomain \to \ksChoiceCoDomain \).
	A relation \( \unlhd \) is called \emph{\( \ksChoiceStructure \)-compatible with \( \ksChoiceFunction \)}, if \( {\unlhd} \subseteq \ksMathcal{E} \times \ksMathcal{E} \) is a \ref{pstl:reflexive}, \ref{pstl:antisymmetric}, \ref{pstl:Suzumura_consistency}, \( \ksChoiceStructure \)-smooth relation on some set \( \ksMathcal{E} \subseteq \ksChoiceCoDomain \), such that the following is satisfied for every \( S \in \ksChoiceDomain \):
	\begin{equation*}\tag{\ref{eq:triangledownLLdefinition}}
		\ksChoiceFunction(S) = \begin{cases}
			E & \ksIF \minOf{S}{{\unlhd}}=\{E\} \\
			K & \ksOtherwise
		\end{cases}
	\end{equation*}
\end{definition}
As one may notice, \( \ksChoiceOrderEF \) in from Definition~\ref{def:BUCcompatible}  is an order some subset \( \ksMathcal{E} \subseteq \ksChoiceDomain \).
The specific set \( \ksMathcal{E} \) we will use will be called the \emph{image of \( \ksChoiceFunction \)}, which is the following:
\begin{equation*}
	\ksChoiceImage{\ksChoiceFunction} = \{\, \ksChoiceFunction(S) \mid S \in \mathbb{S}  \,\}
\end{equation*}
This set satisfied the following property which will be very helpful.
\begin{lemma}\label{lem:BUCimage}
	Let \( \ksChoiceStructure = \tuple{\ksChoiceAlternatives,\ksChoiceDomain,\ksChoiceCoDomain} \) be a restricted choice structure.
If  \( \ksChoiceFunction : \ksChoiceDomain \to \ksChoiceCoDomain \) satisfies \eqref{pstl:SS6E}, then \(  \ksChoiceImage{\ksChoiceFunction} = \{ E \in \ksChoiceCoDomain \mid \ksChoiceFunction(E) = E  \} \).	
\end{lemma}
\begin{proof}
	Observe that for every \( E \in \ksChoiceCoDomain \) there is some \( S \in \mathbb{S} \) with \( \ksChoiceFunction(S)=E \) and \( S \subseteq E \). By setting \( S_1=S_3=E \) and \( S_2=S \) in \eqref{pstl:SS6E},  one obtains that \(  \ksChoiceFunction(S)=\ksChoiceFunction(S_1\cup S_2)=E \) implies that \( \ksChoiceFunction(S_1\cup S_3)=\ksChoiceFunction(E)=E \).
\end{proof}

The encoding of \( \ksChoiceFunction \) works intuitively by saying \( E_1 \) is preferred over \( E_2 \) if there is a set that  witnesses that \( E_1 \) is preferred over \( E_2 \).
A similar encoding scheme has been described by Sen~\cite[Def. 2]{KS_Sen1971}.
\begin{definition}\label{def:encoding_scheme}
	Let \( \ksChoiceStructure = \tuple{\ksChoiceAlternatives,\ksChoiceDomain,\ksChoiceCoDomain} \) be a restricted choice structure and let be a function \( \ksChoiceFunction : \ksChoiceDomain \to \ksChoiceCoDomain \).
	The binary relation \( {\unlhd} \subseteq  \ksChoiceImage{\ksChoiceFunction} \times  \ksChoiceImage{\ksChoiceFunction} \) on \( \ksChoiceImage{\ksChoiceFunction} \) which is given by:
	\begin{align*}        
			E_1 \unlhd E_2 \text{ if }E_1 = K \text{ or there} & \text{ exist some } E\in\ksChoiceDomain \\
		& \text{ such that } E_1\cup E_2 \subseteq E \ksAND \ksChoiceFunction(E) = E_1
	\end{align*}
\end{definition}

\noindent 
We are now present the constructive part of the proof of Theorem~\ref{thm:reptheorem_linearchoiceNonUnion}.
This will especially show that the encoding from Definition~\ref{def:encoding_scheme} is suitable for encode linear choice for arbitrary unrestricted choice structures.

\begin{proposition}\label{prop:compatible}
    Let \( \ksChoiceStructure = \tuple{\ksChoiceAlternatives,\ksChoiceDomain,\ksChoiceCoDomain} \) be a restricted choice structure. 
    For every function \( \triangledown : \ksChoiceDomain \to \ksChoiceCoDomain \) that satisfies \eqref{pstl:SS1}--\eqref{pstl:SS4}, \eqref{pstl:SS5E} and \eqref{pstl:SS6E} there exists a \( K \)-minimal relation \( {\unlhd} \subseteq \textsf{Image}(\triangledown) \times \textsf{Image}(\triangledown) \) that is \( \ksChoiceStructure \)-compatible  with \( \triangledown \).
\end{proposition}
\begin{proof}

    Let \( {\unlhd} \subseteq  \ksChoiceImage{\ksChoiceFunction} \times  \ksChoiceImage{\ksChoiceFunction} \) be the binary relation on \( \ksChoiceImage{\ksChoiceFunction} \) which is given by Definition~\ref{def:encoding_scheme}
    We proof several properties of \( {\unlhd} \):
    \begin{description}    	
    	\item[\normalfont{[}\emph{{\( \min({\unlhd}) = \{K\} \)}.}{]}]
    	Clearly, \( K \unlhd E \) holds for all \( E \in  \textsf{Image}(\triangledown) \).
    	We have to show that \( E \centernot{\unlhd} K \) holds for all \( E \in  \textsf{Image}(\triangledown) \) with \( E\neq K \).
    	Towards a contradiction, suppose there is some \( E \) with \( E \unlhd K \) and \( E\neq K \).
    	Then there must be an \( E' \) with \( E \cup K \subseteq E' \) with \( \ksChoiceFunction(E') = E \).
    	But this is impossible, as we have \( K \subseteq E' \), and thus we have \( \ksChoiceFunction(E')=K \) due to \eqref{pstl:SS2}.

    	\item[\normalfont{[}\emph{{\( {\unlhd} \) is \ref{pstl:reflexive}}.}{]}]
    	We have to show that for every \( E \in \ksChoiceImage{\ksChoiceFunction} \) holds \( E \unlhd E \).
    	By Lemma~\ref{lem:BUCimage} we have \( \ksChoiceFunction(E) = E \), which yields \( \ksChoiceFunction(E \cup E) = E \). Consequently, we also have \( E \unlhd E \).

    	\item[\normalfont{[}\emph{{\( {\unlhd} \) is \ref{pstl:antisymmetric}}.}{]}]
    	We have to show that for all \( E_1,E_2 \in \ksChoiceImage{\ksChoiceFunction} \) with \( E_1 \preceq E_2 \) and \( E_2 \preceq E_1 \) it holds that \( E_1 = E_2 \).
    	Towards a contradiction we assume \( E_1 \neq E_2 \), and let \( E_1 \unlhd E_2 \) and \( E_2 \unlhd E_1 \). 
    	From the last two assumptions we obtain that there are \(  E_{1,2},E_{2,1} \in \ksChoiceDomain \) with \( E_1\cup E_2 \subseteq E_{1,2} \) and \( E_1\cup E_2 \subseteq E_{2,1} \) such that \( \ksChoiceFunction(E_{1,2}) = E_1 \) and \( \ksChoiceFunction(E_{2,1}) = E_2 \).
    	Clearly, we have \( \ksChoiceFunction(E_{1,2}) \subseteq E_{2,1} \) and \( \ksChoiceFunction(E_{2,1}) \subseteq E_{1,2} \), and thus \( \ksChoiceFunction(E_{1,2}) = \ksChoiceFunction(E_{2,1}) \) due to \eqref{pstl:SS3}.
    	We obtain the contradiction of \( E_1 = E_2 \) and \( E_1 \neq E_2 \).
    	Hence, the relation is antisymmetric.

    	\item[\normalfont{[}\emph{{\( {\unlhd} \) is \ref{pstl:Suzumura_consistency}}.}{]}]
    	We have to show that for all \( n\in\mathbb{N}^{+} \) and \( x_0,\ldots,x_n\in X \) holds \( E_0 \unlhd E_1 \), ..., \( E_{n-1} \unlhd E_{n} \) implies that not \( E_{n} \lhd_{\Psi} E_{0} \).
    	Recall that \( E_{n} \lhd_{\Psi} E_{0} \) means that \( E_{n} \unlhd E_{0} \) holds and that \( E_{0} \centernot{\unlhd} E_{n} \) holds.
    	
    	Towards a contradiction, assume that \hyperref[pstl:Suzumura_consistency]{consistency} is violated by \( \unlhd \).
    	This means there exist \( E_0,\ldots,E_n \), for some \(  n\in\mathbb{N}^+ \), such that \( E_{0} \unlhd E_{1} \), ..., \( E_{n-1} \unlhd E_{n} \) and \( E_{n} \lhd E_{0} \). 
    	Because \( \unlhd \) is \ref{pstl:reflexive} and \hyperref[pstl:antisymmetric]{antisymmetric}, \( E_{n} \lhd E_{0} \) implies that \( E_{n} \neq E_{0} \).
    	From the definition of \( {\unlhd} \), we obtain that
    	there are \( E_{0,1},E_{1,2},\ldots,E_{n-1,n} \) and \( E_{n,0} \) with 
    	\begin{align*}
    	E_0 \cup E_1 & \subseteq E_{0,1}\\
    		& \vdots \\
    	E_{n-1} \cup E_n &  \subseteq E_{n-1,n}\\
    	E_n \cup E_0 & \subseteq E_{n,0}
    	\end{align*}
    	 such that the following holds:
    	\begin{align*}
    		\ksChoiceFunction(E_{0,1}) & = E_0 \\
    		& \ \,\vdots \\
    		\ksChoiceFunction(E_{n-1,n}) & = E_{n-1} \\
    		\ksChoiceFunction(E_{n,0}) & = E_n 
    	\end{align*}
    	In summary, we obtain a violation of \eqref{pstl:SS5E}, which is satisfied by \( {\ksChoiceFunction} \) and consequently, we have shown that \( {\unlhd} \) is \ref{pstl:Suzumura_consistency}.

		\item[\normalfont{[}\emph{{If \( \ksChoiceFunction(S) \neq K \), then \( \ksChoiceFunction(S) \in \minOf{S}{\unlhd} \)}.}{]}]
		Towards a contradiction, we assume that \( \ksChoiceFunction(S) \neq K \) and  \( \ksChoiceFunction(S) \notin \minOf{S}{\unlhd} \) hold.
		Let \(  M \in \ksChoiceImage{\ksChoiceFunction} \) with \( M = \ksChoiceFunction(S) \).
		Furthermore, because we have \( \ksChoiceFunction(S) \neq K  \) and because \( \ksChoiceFunction \) satisfied \eqref{pstl:SS1}, we have \( M \subseteq S \).
		From \( \ksChoiceFunction(S) \notin \minOf{S}{\unlhd} \) we obtain that there is some 
		\( N \in \ksChoiceImage{\ksChoiceFunction}  \)
		such that \( N \subseteq S \) and   \( N \unlhd M \).
		From \( N \unlhd M \) we obtain some \( E\in\ksChoiceDomain \) with \( N \cup M \subseteq E \) and \( \ksChoiceFunction(E) = N \).
		We obtain \( \ksChoiceFunction(S) \subseteq E\) and \( \ksChoiceFunction(E) \subseteq S \), and thus, we obtain \( \ksChoiceFunction(S)  = \ksChoiceFunction(E) = M \) from \eqref{pstl:SS3} and \( \ksChoiceFunction(S) = M \). 
		We obtain the  contradiction of having \( M \neq N \) and \( M = N \) at the same time.
		
		\item[\normalfont{[\emph{If \( \ksChoiceFunction(S) \neq K \), then \( \minOf{S}{\unlhd} \) is a singleton set}.]}]
		Towards a contradiction, assume there are sets \( M,N \) with \( M,N \in \min(S,{\unlhd}) \) and \( M \neq N \).
		As shown before, we can safely assume that \( M = \ksChoiceFunction(S) \) holds.
		Because we have \( M,N \in \min(S,{\unlhd}) \), we obtain that \( M \cup N \subseteq S \).
		Thus, from \( \ksChoiceFunction(S) = M \) we obtain \( M \unlhd N \).
		Because \( \unlhd \) is antisymmetric, we obtain \( N \centernot\unlhd M  \) from \( M \neq N \).
		The latter observation is a contradiction to the minimality of \( N \), i.e., \( N \in \min(S,{\unlhd}) \).

        \item[\normalfont{[}\emph{{If \( \triangledown(S) = K \), then \( \minOf{S}{\unlhd}=\{K\} \) or \( \minOf{S}{\unlhd} = \emptyset \)}.}{]}]
        We consider two cases.
        The first case is \( {K \subseteq S} \).
        Recall that we have shown above that \( \min(\unlhd)= \{ K \} \) holds in this case. Hence, we also have \( \minOf{S}{\unlhd}=\{K\} \) whenever \( K \subseteq S \).
        The second case is \( {K \not\subseteq S} \).
        We will show that \( {\minOf{S}{\unlhd} = \emptyset} \).
        Towards a contradiction, we assume that \( \minOf{S}{\unlhd} \) is non-empty, i.e., there is some \( M \in \minOf{S}{\unlhd} \).
        Clearly, by definition, we have that \( M \) satisfies \( M \subseteq S \).
        Moreover, from \( M \in \minOf{S}{\unlhd} \) we obtain \( \triangledown(M) = M  \) and hence, by employing \eqref{pstl:SS4}, we obtain \( \triangledown(S) \subseteq S \) from \( M \subseteq S \). 
        This is a contradiction to \( {K \not\subseteq S} \), because we have \( \triangledown(S) = K \).

    \end{description}
    In summary, we have shown that for every \( S\in\ksChoiceDomain \) holds:
    \begin{equation*}\tag{\#}\label{eq:proof:semrep:eq1}
        \triangledown(S) = \begin{cases}
            E & \ksIF \minOf{S}{{\unlhd}}=\{E\} \\
            K & \ksOtherwise
        \end{cases}
    \end{equation*}
    Moreover, \( {\unlhd} \) is a \ref{pstl:reflexive}, \ref{pstl:antisymmetric}, and \ref{pstl:Suzumura_consistency} relation on \( \textsf{Image}(\triangledown) \).
    Hence, the relation \( {\unlhd} \) is  \( \ksChoiceStructure \)-compatible  with \( \triangledown \).
\end{proof}

One might note that we did not show in Proposition~\ref{prop:compatible} that \( \unlhd \) is transitive or total.
There seems to be no general approach to obtain a transitive and total encoding constructively.
The rationale for \( \unlhd \) being neither transitive nor total is that there is some case where \( \ksChoiceFunction  \) does not provide \enquote{enough} information to order the elements. More specifically, there are situations where \( Q \unlhd M \) and \( Q \unlhd N \) hold, but from \( \ksChoiceFunction \) we cannot obtain any preference on whether \( M \unlhd N \) or \( M \unlhd N \) should hold\footnote{In social choice theory one talks about preferences that are not revealed.}.
One very simple situation is \( M=\{ a \}, N=\{ b \}\), \( Q=\{ a , b \} \) and \( \ksChoiceFunction(Q)=\ksChoiceFunction({M}\cup{N})=Q \). Hence, from that latter choice, we only obtain the local information that  \( Q\unlhd M \) and \( Q\unlhd N \); the choice \( \ksChoiceFunction({M}\cup{N})=Q \) does not tell us if \( M \unlhd N \) or \( M \unlhd N \). In general, we cannot just pick one of \( M \unlhd N \) or \( M \unlhd N \), because that could depend on other choices.
However, using Theorem~\ref{thm:Suzumura_extension} and Proposition~\ref{prop:linear_extension_finite} we can show that non-constructive methods can heal this situation.

		\begin{proposition}\label{prop:BUCleftdirection}
			Assume the Axiom of Choice.
			Let \( \ksChoiceStructure = \tuple{\ksChoiceAlternatives,\ksChoiceDomain,\ksChoiceCoDomain} \) be a restricted choice structure. 
			For every function \( \ksChoiceFunction : \ksChoiceDomain \to \ksChoiceCoDomain \) that satisfies \eqref{pstl:SS1}--\eqref{pstl:SS4}, \eqref{pstl:SS5E} and \eqref{pstl:SS6E} for some fixed \( K\in\ksChoiceCoDomain \) there is a \( K \)-minimal linear order \( {\lll} \subseteq \ksMathcal{E} \times \ksMathcal{E} \) on some set \( \ksMathcal{E} \subseteq \ksChoiceCoDomain \) that is \( \ksChoiceStructure \)-compatible with \( \ksChoiceFunction \) and \( R \)-smooth.
		\end{proposition}
\begin{proof}
	Assume that \( \ksChoiceFunction \) satisfies \eqref{pstl:SS1}--\eqref{pstl:SS4}, \eqref{pstl:SS5E} and \eqref{pstl:SS6E} for some fixed \( K\in\ksChoiceCoDomain \).
	From Proposition~\ref{prop:compatible} we obtain a relation \( {\unlhd} \subseteq \ksChoiceImage{\ksChoiceFunction} \times \ksChoiceImage{\ksChoiceFunction} \) that is \( \ksChoiceStructure \)-compatible  with \( \ksChoiceFunction \).
	Due to Suzumura's theorem, Theorem~\ref{thm:Suzumura_extension}, there exists a total preorder \( {\sqsubseteq} \) on \( \ksChoiceImage{\ksChoiceFunction} \) such that \( {\unlhd} \subseteq {\sqsubseteq} \) and \( {\lhd} \subseteq {\sqsubset} \) hold.
	We will show that for all \( S\in\ksChoiceDomain \) holds:
	\begin{equation*}
		\minOf{S}{{\sqsubseteq}} \quad=\quad \minOf{S}{{\unlhd}}
	\end{equation*}
	We show that the equivalence holds by showing two set inclusions.
	\begin{description}                
		\item[\normalfont{[}\emph{{\( \minOf{S}{{\sqsubseteq}} \subseteq \minOf{S}{{\unlhd}}  \)}.}{]}]
		Assume the contrary. Let \( M \) be an element from \( \ksChoiceImage{\ksChoiceFunction} \) such that \( M \subseteq S \) and \( M \in \minOf{S}{{\sqsubseteq}} \) and \( M \notin \minOf{S}{{\unlhd}} \) hold.
		Because \( M \notin \minOf{S}{{\unlhd}}  \) holds, there is some \( Q \in \ksChoiceImage{\ksChoiceFunction} \) with \( Q \subseteq S \) and \( Q \lhd M \) and \( M \neq Q \).
		Then, because we have \( {\lhd} \subseteq {\sqsubset} \), we also have \( Q \sqsubset M \). 
		We obtain a contradiction, because \( Q \sqsubset M \) and \( Q \subseteq S \) contradict that \( M \) is a minimal element, i.e., that \( M \in \minOf{S}{{\sqsubseteq}} \) holds.
		
		\item[\normalfont{[}\emph{{\( \minOf{S}{{\unlhd}}  \subseteq   \minOf{S}{{\sqsubseteq}} \)}.}{]}]
		Assume the contrary.
		Let \( M \) be an element from \( \ksChoiceImage{\ksChoiceFunction} \) such that \( M \subseteq S \) and \( M \in \minOf{S}{{\unlhd}} \) and \( M \notin \minOf{S}{{\sqsubseteq}} \) hold.
		In Proposition~\ref{prop:compatible}, it is shown that \( M \in \minOf{S}{{\unlhd}} \) implies that \( \minOf{S}{{\unlhd}}  = \{ M \} \).
		Hence, because \( \minOf{S}{{\sqsubseteq}} \subseteq \minOf{S}{{\unlhd}}  \) holds, we obtain that \( \minOf{S}{{\sqsubseteq}}  = \emptyset \).
		The latter implies that there is some \( Q \in \ksChoiceImage{\ksChoiceFunction} \) with \( Q \subseteq S \) and \( Q \sqsubset M \) and \( M \neq Q \).
		Because of \( \minOf{S}{{\unlhd}}  = \{ M \} \) and \( {\unlhd} \) is \ref{pstl:antisymmetric}, there are two cases: \( M \unlhd Q \) or \( M,Q \) are \( \unlhd \)-incomparable.
		Clearly, the case of \( M \unlhd Q \) is impossible due to \( Q \sqsubset M \).     
		Leaving that \( M,Q \) are \( \unlhd \)-incomparable as the only option. 
		Now, obtain \( \ksChoiceFunction(S)=M \) from \( \minOf{S}{{\unlhd}} = \{ M \} \) and the \( \ksChoiceStructure \)-compatibility of \( \unlhd \) with \( \ksChoiceFunction \).
		Because we have \( Q \subseteq S \) and \( M \subseteq S \), we also have \( M \cup Q \subseteq S \). 
		Hence, by inspecting the construction of \( \unlhd \) (see Definition~\ref{def:encoding_scheme}), we obtain \( M \unlhd Q \) from \( \ksChoiceFunction(S)=M \).
		The latter is a contradiction to the observation that \( M,Q \) are \( \unlhd \)-incomparable. 
	\end{description}
	Next, recall that Proposition~\ref{prop:linear_extension_finite} guarantees that there exists a linear order \( {\lll} \) on \( \ksChoiceImage{\ksChoiceFunction} \) such that \( {\sqsubset} \subseteq {\lll} \).
	We will show that \( \lll \) satisfies several properties.
	\begin{description}
		\item[\normalfont{[}\emph{{\( \lll \) is compatible with \( \ksChoiceFunction\)}.}{]}]
		We have to show that the following holds:
		\begin{equation*}\tag{\( \star \)}\label{eq:proof:semrep:eq2}
			\ksChoiceFunction(S) = \begin{cases}
				\minsOf{S}{{\lll}} & \ksIF \minsOf{S}{{\lll}} \text{ is defined} \\
				K & \ksOtherwise
			\end{cases}
		\end{equation*}
		First, we show that \( \minOf{S}{\unlhd} = \minOf{S}{{\lll}}\) holds.
		Recall that we have shown above that \( \minOf{S}{\unlhd} = \minOf{S}{{\sqsubseteq}}\) holds.
		Thus, from \( {\sqsubset} \subseteq {\lll} \), we obtain that \( \minOf{S}{\unlhd} \subseteq \minOf{S}{\lll} \).
		Consequently, because \( \lll \) is a linear order, we have \( |\minOf{S}{\lll}| \leq 1 \), and thus, \( \minOf{S}{\unlhd} = \minOf{S}{\lll} \) whenever \( \minOf{S}{\unlhd} \neq \emptyset \).
		It remains to show that \( \minOf{S}{\unlhd} = \emptyset \) implies \( \minOf{S}{\lll} = \emptyset \).
		Towards a contradiction, let \( M \in \minOf{S}{\lll}  \), but \( \minOf{S}{\unlhd} = \emptyset \).
		Because \( M \notin \minOf{S}{{\unlhd}}  \) holds, there is some \( Q \in \ksChoiceImage{\ksChoiceFunction} \) with \( Q \subseteq S \) and \( Q \lhd M \) and \( M \neq Q \).
		Then, because we have \( {\lhd} \subseteq  {\sqsubset} \subseteq {\lll} \), we also have \( Q \lll M \). 
		We obtain a contradiction, because \( Q \lll M \) and \( Q \subseteq S \) contradict that \( M \) is a minimal element, i.e., that \( M \in \minOf{S}{{\unlhd}} \) holds.
		This completes the proofs of \( \minOf{S}{\unlhd} = \minOf{S}{{\lll}}\).
		Now observe that \( \minsOf{S}{{\lll}} \) is defined if and only if \( |\minOf{S}{{\lll}}| = 1\).
		Moreover, if \( \minsOf{S}{{\lll}} \) is defined, then \( \minsOf{S}{{\lll}} =M  \) with \( \minOf{S}{{\lll}} =\{ M \} \).
		We obtain \eqref{eq:proof:semrep:eq2} from \eqref{eq:proof:semrep:eq1}.
		
		\item[\normalfont{[}\emph{{\( \mins(\lll) = K \)}.}{]}]
		We have shown above that \( \minOf{S}{\unlhd} = \minOf{S}{{\lll}}\) holds.
		Furthermore, we also have shown that \( \min(\unlhd) = \{ K \} \) holds.
		By combining these statements, we obtain \( \mins(\lll) = K \) by employing the definition of \( \mins \).
	\end{description}
	In summary, we obtain that \(  {\lll} \) is \( \ksChoiceStructure \)-compatible with \( \ksChoiceFunction \), and \( K \)-minimal
	and \( R \)-smooth.
	\qedhere
\end{proof}

The propositions above give rise to Theorem~\ref{thm:reptheorem_linearchoiceNonUnion}.

{\medskip\noindent\textbf{Theorem~\ref{thm:reptheorem_linearchoiceNonUnion} {(Representation Theorem)}.}~\textit{%
        Assume the Axiom of Choice.
        Let \( \ksChoiceStructure = \tuple{\ksChoiceAlternatives,\ksChoiceDomain,\ksChoiceCoDomain} \) be a restricted choice structure and let \( K \in \ksChoiceCoDomain \). 
        A function \( \triangledown : \ksChoiceDomain \to \ksChoiceCoDomain \) is a \( K \)-minimal linear choice function for \( \ksChoiceStructure \) if and only if the axioms \eqref{pstl:SS0a}--\eqref{pstl:SS4}, \eqref{pstl:SS5E} and \eqref{pstl:SS6E} are satisfied.
}}
\begin{proof}
We show each direction independently.
{[}\emph{\enquote{\( \Rightarrow\)}}{]} 
Corollary~\ref{prop:rightdirection_unionclosed} yields the left-to-right direction.
{[}\emph{\enquote{\( \Leftarrow  \)}}{]}
By employ Proposition~\ref{prop:BUCleftdirection}, we obtain from the satisfaction of \eqref{pstl:SS1}--\eqref{pstl:SS6} by \( \triangledown \) that there is a linear order \( {\lll} \subseteq \ksMathcal{E} \times \ksMathcal{E} \) on some set \( \ksMathcal{E} \subseteq \ksChoiceCoDomain \) that is \( \ksChoiceStructure \)-compatible with \( \triangledown \), and \( K \)-minimal and \( R \)-smooth.
Because \( \triangledown \) satisfies \eqref{pstl:SS0a}, we have for all \( S \in \ksChoiceDomain \) that \( \triangledown(S) \subseteq S \) holds if and only if there is some \( E \in \ksChoiceCoDomain \) such that \( E \subseteq S \).
This implies we have for all \( S \in \ksChoiceDomain \) that \( \minsOf{S}{\lll}  \subseteq S \) if and only if there is some \( E \in \ksChoiceCoDomain \) such that \( E \subseteq S \).
We extend \( \lll \) to a linear order \( {\ll} \subseteq \ksChoiceCoDomain \times \ksChoiceCoDomain  \) on \( \ksChoiceCoDomain \) as follows:
\begin{equation*}
	S_1 \ll S_2 \text{ if }  \left(\ S_1  \lll S_2 \ \right) \ksOR \left(\ S_1 \in \ksMathcal{E} \ksAND s_2 \in \ksChoiceCoDomain\setminus\ksMathcal{E} \ \right) \ksOR \left(\ S_1  \Subset S_2 \ \right)
\end{equation*}
whereby \( {\Subset} \subseteq (\ksChoiceCoDomain\setminus\ksMathcal{E}) \times (\ksChoiceCoDomain\setminus\ksMathcal{E})  \) is some arbitrary linear order on \( \ksChoiceCoDomain\setminus\ksMathcal{E} \), whose existence is guaranteed by the Axiom of Choice.
Note that \( \ll \) is \( \ksChoiceStructure \)-smooth, \( K \)-minimal and we have \( \minsOf{S}{\ll} = \minsOf{S}{\lll} \) for all \( S \in \ksChoiceDomain \).
Inspecting Definition~\ref{def:linear_choice_function} reveals that we have \( {\triangledown} = {\triangledown_{\ll}^{K}} \) for \( \ll \) because \( \lll \) is \( \ksChoiceStructure \)-compatible with \( \triangledown \).
Thus, we have that \( \triangledown \) is a \( K \)-minimal linear choice function.
\end{proof}

We show that for union-closed restricted choice structure, satisfaction of \eqref{pstl:SS5E} and \eqref{pstl:SS6E} coincides with satisfaction of \eqref{pstl:SS5} and \eqref{pstl:SS6}. 
\begin{proposition}\label{prop:EandNonE}
    Assume the Axiom of Choice.
    Let \( \ksChoiceStructure = \tuple{\ksChoiceAlternatives,\ksChoiceDomain,\ksChoiceCoDomain} \) be a union-closed restricted choice structure and let \( K \in \ksChoiceCoDomain \) . 
    For every function \( \triangledown : \ksChoiceDomain \to \ksChoiceCoDomain \), the following statements are equivalent:
    \begin{enumerate}[(a)]
        \item \( \triangledown \) satisfies \eqref{pstl:SS5} and \eqref{pstl:SS6} 
        \item \( \triangledown \) satisfies \eqref{pstl:SS5E} and \eqref{pstl:SS6E} 
    \end{enumerate}
\end{proposition}
\begin{proof}
Because \( \ksChoiceStructure \) is union-closed, we have \( S_1 \cup S_2 \in \ksChoiceDomain \) and \( S_1 \cup S_3 \in \ksChoiceDomain \) for all \( S_1,S_2,S_3 \in \ksChoiceDomain \).
Consequently, \eqref{pstl:SS6E} and \eqref{pstl:SS6} are the same postulate for union-closed restricted choice structures.

Next, we observe that  \eqref{pstl:SS5E} implies \eqref{pstl:SS5}.
Because \( \ksChoiceStructure \) is union-closed, we have \( S_n \cup S_0 \in \ksChoiceDomain \) and \( S_0 \cup S_{1} \in \ksChoiceDomain \), ..., \( S_{n-1} \cup S_{n} \in \ksChoiceDomain \) for all \( S_0,\ldots,S_n \in \ksChoiceDomain \). 
Consequently, \eqref{pstl:SS5} is a special case of \eqref{pstl:SS5E} and thus, \eqref{pstl:SS5E} implies \eqref{pstl:SS5}.

We show that satisfaction of \eqref{pstl:SS5} and \eqref{pstl:SS6} imply satisfaction of \eqref{pstl:SS5E}.
Let \(  S_{i}\cup S_{i+1} \subseteq S_{i,i+1} \) and \( \triangledown(S_{i,i+1})  = S_{i} \) for \( 0\leq i\leq n \) and let \( S_n \cup S_0 \subseteq S_{n,0} \) and \( S_{n} \neq S_{0} \).
We show that \( \triangledown(S_{n,0})  \neq S_{n} \) holds.
The proof is by contradiction, i.e., we assume that \( \triangledown(S_{n,0})  = S_{n} \) holds.
By employing \eqref{pstl:SS6},  we obtain \( \triangledown(S_{i+1}  \cup S_{i} ) = S_{i} \) from  \( \triangledown(S_{i,i+1}) = \triangledown(S_{i+1}  \cup S_{i,i+1} ) = S_{i} \) for \( 0\leq i\leq n \). 
Clearly, we also obtain \( \triangledown(S_{n}  \cup S_{0} ) = S_{n} \) analogously.
This is a contradiction, because from \eqref{pstl:SS5} we obtain that \( \triangledown(S_n \cup S_0)  \neq S_{n} \) holds.
For that we employ that \( S_{n} \neq S_{0} \) and \( \triangledown(S_{i+1}  \cup S_{i} ) = S_{i} \) for \( 0\leq i\leq n \) holds.
\end{proof}
\noindent Conjoin Theorem~\ref{thm:reptheorem_linearchoiceNonUnion} and Proposition~\ref{prop:EandNonE} yields Theorem~\ref{thm:reptheorem_linearchoice} immediately.

{\medskip\noindent\textbf{Theorem~\ref{thm:reptheorem_linearchoice}.}~\textit{%
		Assume the Axiom of Choice.
		Let \( \ksChoiceStructure = \tuple{\ksChoiceAlternatives,\ksChoiceDomain,\ksChoiceCoDomain} \) be a union-closed restricted choice structure and let \( K \in \ksChoiceCoDomain \). 
		A function \( \triangledown : \ksChoiceDomain \to \ksChoiceCoDomain \) is a \( K \)-minimal linear choice function for \( \ksChoiceStructure \) if and only if the axioms \eqref{pstl:SS0a}--\eqref{pstl:SS6} are satisfied.
}}

\medskip
One can show that in the case of union-closed restricted choice structure, an easier encoding scheme that is different from Definition~\ref{def:encoding_scheme} is also sufficient. The encoding schemes, which is known from, e.g.,  theory change \cite{KS_KatsunoMendelzon1991,KS_FalakhRudolphSauerwald2022}, is the following:
\begin{equation*}
	E_1 \unlhd E_2 \text{ if } \ksChoiceFunction(E_1\cup E_2) = E_1 \tag{\( \Box \)}\label{eq:Box}
\end{equation*}
For the proof of Theorem~\ref{thm:reptheorem_linearchoiceNonUnion}, respectively Proposition~\ref{prop:compatible}, we cannot use such an encoding scheme.
This is because \( \ksChoiceDomain \) is in general not union-closed and thus one cannot safely assume that \( E_1 \cup E_2 \in \ksChoiceDomain \) holds.
More precisely, when one would only consider cases where \( E_1 \cup E_2 \) from \eqref{eq:Box} exist, the order \( \unlhd \) via \eqref{eq:Box} would not fully encode \( \ksChoiceFunction \), i.e., \( \minOf{S}{\unlhd} \) would differ from \( \ksChoiceFunction(S) \) for some \( S \). \end{document}